\documentclass{sbc2023}%

\usepackage{graphicx}
\usepackage[utf8]{inputenc}
\usepackage[misc,geometry]{ifsym} 
\usepackage[T1]{fontenc}
\usepackage[utf8]{inputenc}
\usepackage{fontawesome}
\usepackage{orcidlink} 
\usepackage{color}
\usepackage{aas_macros}
\usepackage[bottom]{footmisc}
\usepackage{supertabular}
\usepackage{afterpage}
\usepackage{xurl}
\usepackage{hyperref} 

\usepackage{pifont}
\usepackage{multicol}
\usepackage{multirow}
\usepackage{booktabs}
\usepackage{textcomp,gensymb}
\usepackage{amssymb}
\usepackage{adjustbox}
\usepackage{mdframed}
\usepackage{enumitem}
\setlist[itemize]{noitemsep, before={\vspace*{-0.1cm}}}

\newcommand{\PeriodPunct}{\mathpunct{\raisebox{0.5ex}{.}}}

\setcitestyle{square}

\definecolor{orcidlogo}{rgb}{0.37,0.48,0.13}
\definecolor{unilogo}{rgb}{0.16, 0.26, 0.58}
\definecolor{maillogo}{rgb}{0.58, 0.16, 0.26}
\definecolor{darkblue}{rgb}{0.0,0.0,0.0}
\hypersetup{colorlinks,breaklinks,
            linkcolor=darkblue,urlcolor=darkblue,
            anchorcolor=darkblue,citecolor=darkblue}

\definecolor{Gray}{gray}{0.9}

\hypersetup{colorlinks,citecolor=blue,linkcolor=blue,urlcolor=blue}

\title{Building High-Quality Datasets for Portuguese LLMs: From Common Crawl Snapshots to Industrial-Grade Corpora}

\author[Almeida et al. 2025]{

\affil{\textbf{Thales Sales Almeida}~\orcidlink{0009-0006-9568-9331}[~\textbf{Institute of Computing, University of Campinas, Maritaca AI}~|\href{mailto:t224732@dac.unicamp.br}{~\textbf{\textit{t224732@dac.unicamp.br}}}~]}

\affil{\textbf{Rodrigo Nogueira}~\orcidlink{0000-0002-2600-6035}[~\textbf{Maritaca AI}~|\href{mailto:rfn@unicamp.br}{~\textbf{\textit{rodrigo@maritaca.ai}}}~]}

\affil{\textbf{Helio Pedrini}~\orcidlink{0000-0003-0125-630X}~~\textcolor{blue}{\faEnvelopeO}~[~\textbf{Institute of Computing, University of Campinas}~|\href{mailto:helio@ic.unicamp.br}{~\textbf{\textit{helio@ic.unicamp.br}}}~]}
}

\begin{document}

\tolerance=999
\sloppy

\begin{frontmatter}
\maketitle

\begin{mail}
Institute of Computing, University of Campinas, Av. Albert Einstein, 1251, Campinas, SP, 13083-852, Brazil \\
Maritaca AI, Campinas, SP, Brazil
\end{mail}





\begin{abstract}
\textbf{Abstract.~}
\noindent The performance of large language models (LLMs) is deeply influenced by the quality and composition of their training data. While much of the existing work has centered on English, there remains a gap in understanding how to construct effective training corpora for other languages. We explore scalable methods for building web-based corpora for LLMs. We apply them to build a new 120B token corpus in Portuguese that achieves competitive results to an industrial-grade corpus. Using a continual pretraining setup, we study how different data selection and preprocessing strategies affect LLM performance when transitioning a model originally trained in English to another language. Our findings demonstrate the value of language-specific filtering pipelines, including classifiers for education, science, technology, engineering, and mathematics (STEM), as well as toxic content. We show that adapting a model to the target language leads to performance improvements, reinforcing the importance of high-quality, language-specific data. While our case study focuses on Portuguese, our methods are applicable to other languages, offering insights for multilingual LLM development.
\end{abstract}

\begin{keywords}
Large Language Models, Large Pretraining Corpus, Multilingual Models
\end{keywords}

\end{frontmatter}

\section{Introduction}

The success of large language models (LLMs) largely depends on the data they are trained on. These models are built on massive collections of text, primarily sourced from the web, and their capabilities are shaped by the quality and quantity of this data~\citep{liu2024datasets,dubey2024llama3,fineweb}. As LLMs scale in size and complexity, selecting and curating training data has emerged as a central challenge~\citep{fineweb,penedo2023refinedweb,kreutzer2022quality}. Decisions about which sources to include -- and how to filter, balance, and structure them -- play a crucial role in shaping model performance and generalization~\citep{goupher_massiveweb}.


Verifying the quality of training data is a complex but essential component of developing effective LLMs. Typically, this involves curating a dataset, training a model, and evaluating the results against established benchmarks. While there has been substantial progress in data quality assessment, most of it has focused on English datasets~\citep{gao2020pile,c4_intro,penedo2023refinedweb}, with Chinese emerging as a secondary focus~\citep{longpre2024bridgingprovenancegap}. This leaves a considerable gap in our understanding of how to develop and optimize LLMs for other languages~\citep{almeida2025tiebe,moayeri2024worldbench}.

Multilingual initiatives such as mC4~\citep{mc4}, BLOOM~\citep{bloom}, MADLAD~\citep{madlad}, and CulturaX~\citep{culturax} have attempted to bridge this gap. However, their aim to support over a hundred languages simultaneously often results in limited per-language representation~\citep{bloom} and inconsistent data quality across languages~\citep{kreutzer2022quality}. 

In this work, we focus on Portuguese, a language that remains underexplored in LLM research. Despite being the fifth most spoken language worldwide, Portuguese still has a limited amount of large-scale pretraining corpus. To address this, we investigate methods for constructing high-quality Portuguese corpora through large-scale web crawls from Common Crawl, applying various preprocessing and filtering techniques. Though we focus on Portuguese, we believe our findings and pipeline can be replicated for other languages. 

Given the high computational cost of training LLMs from scratch, our approach in this work emphasizes continual pretraining from open-weight models. We study how different data selection methods and preprocessing strategies influence performance when transitioning a model extensively trained in English~\citep{zhang2024tinyllama} to a new language like Portuguese. This setup enables us to evaluate the effects of data quality on model adaptation~\citep{gogoulou2024languageshift}.

Our main contributions are as follows:

\begin{itemize}

\item We introduce ClassiCC-PT, a large web corpus in Portuguese with 120B tokens. We demonstrate that our dataset is competitive with other industry-grade corpora of the same size~\citep{overwijk2022clueweb22}.

\item We analyze how various data processing techniques and selection strategies influence the performance of large language models (LLMs) during continued pretraining.

\item We highlight the advantages of developing language-specific classifiers and outline a method for creating new ones. Additionally, we introduce three BERT-based classifiers designed for education, STEM, and toxic content.

\item We reinforce the evidence that training the LLM in the target language yields higher performance than further training in English.

\item We show that substantial performance gains can be achieved through continual pretraining in a new language (Portuguese), even when starting from a base model originally trained exclusively in another language (English).

\end{itemize}

\section{Related Work}

This section surveys prior work across several key areas relevant to our study: language specialization, the composition and curation of pretraining corpora, methodological advances in data selection, the cost-efficient practice of continued pretraining, and the persistent disparities in corpus representation that affect performance across different regions and languages.

\subsection{Language Specialization}

Research in the field of LLMs has increasingly focused on specialization to improve performance in specific domains such as finance~\citep{bloomberggpt}, law~\citep{saulm}, mathematics~\citep{lewkowycz2022solving,azerbayev2024llemma}, and programming~\citep{magicoder, codellama}. These specialized models are trained on domain-specific datasets to enhance their expertise, resulting in superior task performance within those fields. For instance, models trained on legal documents can better understand complex legal jargon, while those exposed to vast amounts of code can better assist with programming tasks and code generation. The success of these specialized models underscores the importance of targeted training data in achieving better results.

A parallel line of research has explored language specialization, that is, adapting or training models to excel in a particular language rather than a topical domain. While most LLM development has focused on English, due to its abundant resources and research attention, several studies have shown that specializing models in a target language can yield substantial performance gains, particularly on benchmarks in that language~\citep{gogoulou2024languageshift,sabia2,sabia3,typhoon,qwen,eurollm,seallm}. This approach acknowledges that language can be a critical axis of specialization, allowing models to better capture linguistic nuances and cultural context that general-purpose, English-centric models may overlook.

Several recent studies have focused on developing specialized language models for Portuguese at varying scales~\citep{bertimbau,gervasio,albertina,gloria}. The Sabia models (7B and 65B) are based on Llama 7B and 65B, respectively, and are further trained on 10 billion tokens from ClueWeb-A, resulting in substantial performance gains. Tucano~\cite{correa2024tucano} introduces models with up to 2.4 billion parameters, leveraging Gigaverbo, a mixture of datasets refined using custom classifiers. This work presents Curi\'o 1.1B, a model trained on 120 billion unique tokens. Curi\'o operates at a scale comparable to Tucano but is initialized from TinyLlama 1T~\citep{zhang2024tinyllama} rather than being trained from scratch. In our evaluations, Curi\'o 1.1B outperforms Tucano 1.1B, which we attribute to the advantages of continual pretraining on top of TinyLlama.


\subsection{Pretraining Corpus}

Data quality is a driving factor in the quality of machine learning models in general. Large language models are no different: they are typically trained on massive corpora that include documents of varying quality, which are naturally hard to clean and handle.

A popular source for LM training is the Common Crawl project\footnote{\url{https://commoncrawl.org/}}, which collects documents from the web. The Common Crawl dataset is enormous but comes with the drawback of being very variable regarding data quality. Multiple datasets build upon Common Crawl, mainly trying to enhance data quality by selecting documents of higher quality. The C4 and mC4~\citep{c4_intro,c4Documenting,mc4} datasets aim to achieve this by applying statistical filters, for example, the removal of documents from an URL list of problematic sites, such as known sources of pornographic and malware content, or removing documents that contain certain bad words, such as slurs and vulgar language. OSCAR~\citep{oscar_intro} tries to enhance data quality by adding more metadata to each document, labeling the language, and removing clearly noisy documents, such as ones with less than one sentence or with a high percentage of text containing unknown words. Meanwhile, The Pile~\citep{gao2020pile} selects a subset of HTML pages from the Common Crawl, selected using the jusText algorithm~\citep{justext}, and adds a variety of smaller, higher-quality corpora such as Wikipedia and BookCorpus. The idea is that the construction of a diverse corpus would lead to a better LM~\citep{mixture_datasets_1}.

Data curation studies often build upon previous work. In this context, CulturaX~\citep{culturax} was introduced, merging documents from both mC4 and OSCAR and applying further document selection. Notably, it employs metric-based filtering, a technique initially introduced in the BLOOM pipeline~\citep{bloom}. This kind of filtering analyzes different metrics of the documents, such as stop word ratio and perplexity score of a 5-gram Kneser-Ney language model~\citep{heafield2011kenlm}, and removes documents that present outlier values. Furthermore, CulturaX performs a simple version of document refinement: they edit the document by removing certain portions that are likely to be noisy text from the HTML extraction, such as footer information.

MADLAD-400~\citep{madlad} represents a recent effort to provide an extensive multilingual corpus encompassing more than 400 languages. Originating from Common Crawl documents, the methodology involves an iterative document inspection and the development of statistical filters to exclude low-quality documents. An interesting aspect of the MADLAD-400 dataset is its sentence-level classification, which considers factors such as whether a sentence is in a different language than the rest of the document or if it is excessively long or short. A document is filtered out if a certain percentage of its sentences fail to meet the criteria. The work highlights the complexities of formulating heuristics for document selection across various languages, often necessitating the creation of language-specific rules.

Gigaverbo~\citep{correa2024tucano} is a recent large Portuguese corpus with more than 100M documents; it aggregates multiple Portuguese subsets from previously existing datasets and then further filters documents using a general-purpose quality classifier developed by Gigaverbo authors. The corpus was used in the train of the Tucano models, Portuguese LLMs up to 2.4B parameters trained from scratch.

Finally, FineWeb~\citep{fineweb} was an English dataset that, similar to our work, proposed a pipeline for creating high-quality datasets starting from Common Crawl snapshots; they also demonstrated promising results by using a neural network to score documents by their educational value. Later, FineWeb2~\citep{penedo2024fineweb-2} was released as a multilingual expansion of the FineWeb pipeline. However, despite covering more than one thousand languages, there was a reduced effort in some filtering steps, such as the absence of educational scores for non-English documents.

\subsection{Data Curation Methods}

A common trend leveraged by these large-scale corpora is the usage of rule-based filters that draw insights from both document structure and metadata, including attributes such as source URLs. This approach has garnered popularity due to its efficiency, a critical factor when dealing with such substantial text corpora~\citep{goupher_massiveweb,oscar_intro}. However, limited research exists to study the exact effects of these filters on the resulting LM~\citep{penedo2023refinedweb}.

A different perspective for data curation was proposed by~\citet{marion2023less}, consisting of isolating a portion of the training corpus that, when ``pruned'', would lead to improvements in the resulting LM. This method is highly inspired by data pruning used in the computer vision field~\citep{pruning_1,pruning_2}. In particular, the authors find that ranking documents by their perplexity and then pruning the documents with the highest perplexity leads to a better final model in terms of perplexity. However, there was not an extensive study of the impacts of such a technique on downstream tasks.

Within the domain of machine learning-based selection methods, the work developed by~\citet{gpt3_base}, which employs a logistic regression classifier for training data curation, stands out as a notable example. Building upon this, research has evolved to explore the use of the LM itself to filter harmful content, adhering to human-provided guidelines~\citep{bai2022constitutional}. These advances have recently led to the Qwen model~\citep{qwen} demonstrating impressive results when trained on a corpus curated through a combination of rule-based and neural-based methods. However, the details of such neural-based methods were not made public. In this context, our work provide studies for a neural-based method of selection that can be guided by instructions in natural language.

\subsection{Continued Pretraining}

Recent frontier models train hundreds of billions of parameters on tens of trillions of tokens. Naturally, such training runs have a high computational cost. In this scenario, continued pretraining emerges as a strategy for improving results with fewer computation resources.

The basic premise is to start from an existing model and then perform another pretraining on top of such a model using a more specialized dataset, like a dataset in a specific language or that focuses on a specific domain. By starting from the original model weights, the assumption is that part of the original model knowledge is retained while adding new information on the domain of interest via the new pretraining. This approach has gained popularity, being employed in specializing models in code~\citep{codellama}, medicine~\citep{biomistral,chen2023meditron} and law~\citep{saulm, bloomberggpt}.

Such specialization also occurs at language level. In special in the Portuguese context, Sabiá~\citep{sabia} showed that is possible to have considerable gains with a fraction of the computational resources using such strategy. 

In this work, we will perform most of our training runs in a continued pretraining setting since this is a cost-effective way to achieve high-performing models in non-English languages. We will also explore and measure the benefits of continued pretraining by comparing our runs with training runs starting from scratch.

\subsection{Disparities in Corpus Representation}

Recent studies demonstrate that LLMs present higher knowledge gaps depending on the regional context. For instance, WorldBench~\cite {moayeri2024worldbench} is a benchmark that leverages public economic data from the World Trade Center to measure how well LLMs know such data for different countries. It showed that LLMs show a much higher error rate when asked about countries with lower economic status. Additionally, Timely Events Benchmark (TiEBe)~\citep{almeida2025tiebe} is a benchmark that uses Wikipedia retrospective pages to recover noticeable events from various countries. These events are then used to create QA pairs, which are later used to evaluate LLMs factual recall. They also observed a high disparity in LLM performance when comparing factual recall for events originating in the United States, when compared to other regions such as France or Brazil.

These knowledge gaps are likely related to the small representation of such regions in pretraining corpora. \citet{longpre2024bridgingprovenancegap} performed a large audit of text, video, and speech datasets and showed that the representation of South American-originated data is less than 0.2\% on average. They also pointed out that this proportion does not seem to be improving in recent years.

\section{Methodology}

In this section, we describe the methodology used for our experiments, including the dataset creation, model training, and evaluation.

%


\begin{table*}[!htb]
\setlength{\tabcolsep}{3.6mm}
\centering
\caption{Information on the document and token count in the final dataset, as well as the percentage of  Portuguese pages for each of the used crawls in the ClassiCC-PT corpus.}
\label{tab:dataset_docs_and_tokens}
\begin{tabular}{lcccc}
\toprule 
Source    & CC-2021-31 & CC-2021-39 & CC-2022-40 & ClassiCC-PT \\
\midrule
Documents & 62M        & 33M        & 21M        & 116M        \\
Tokens    & 66B        & 36B        & 24B        & 126B  \\  
\midrule
PT percentage in crawl    & 2.2\%        & 2.1\%        & 1.15\%        & -  \\  
\bottomrule
\end{tabular}
\end{table*}

\subsection{Dataset}

The methodology adopted in this work focuses on designing and evaluating a pipeline for constructing non-English text corpora, with Portuguese being our case study. The initial phase involves processing data obtained from a Common Crawl (CC) snapshot, followed by filtering and recovering relevant web pages, specifically those in Portuguese. The next stage encompasses document processing and classification, which includes text cleaning, deduplication, and content categorization to effectively structure the dataset.

\begin{figure}[!htb]
\centering
\includegraphics[width=0.99\linewidth]{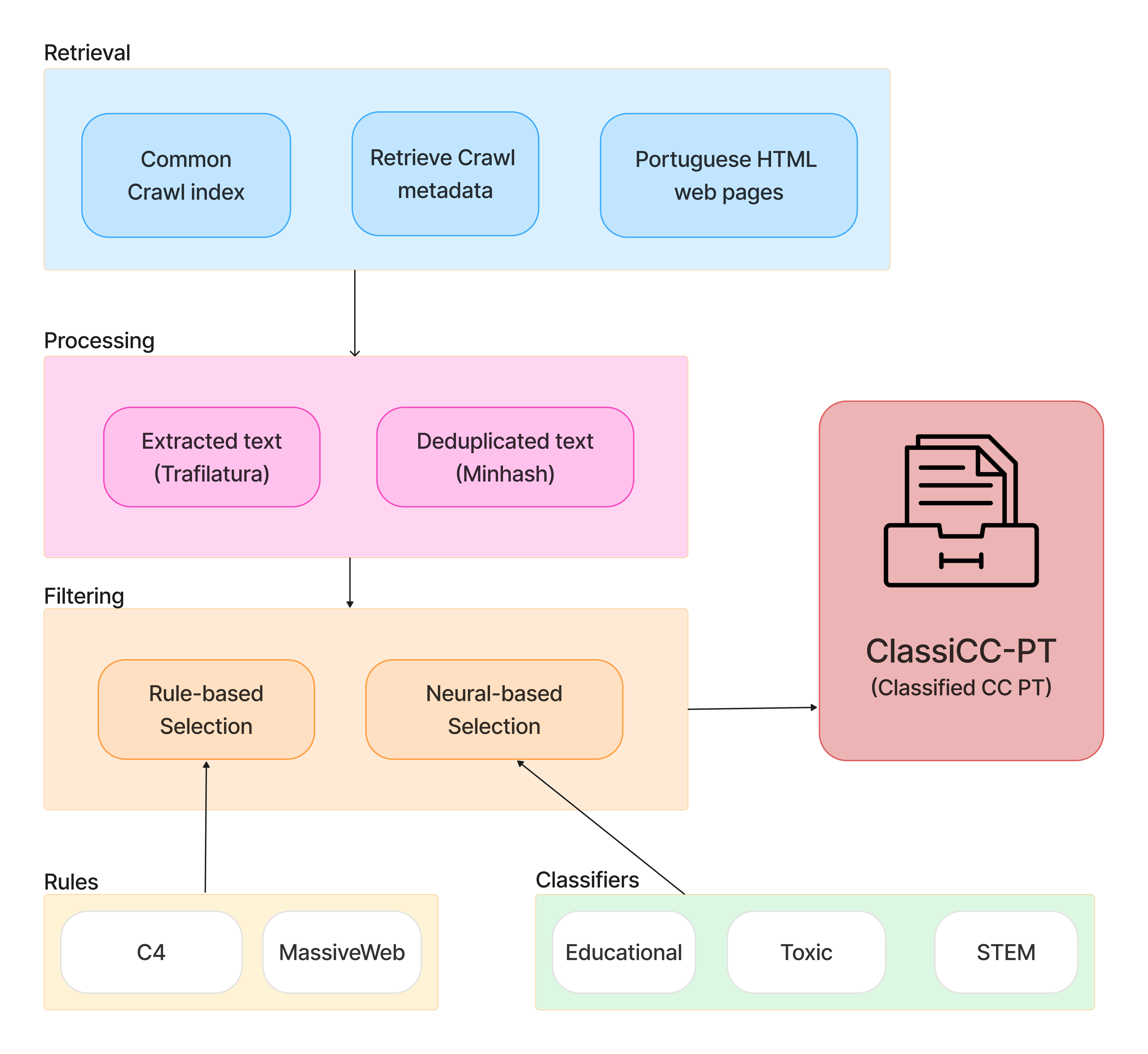}
\caption{The creation process of ClassiCC-PT.}
\label{fig:process}
\end{figure}

The resulting corpus, referred to as Classified CC-PT (ClassiCC-PT)\footnote{\url{https://huggingface.co/datasets/ClassiCC-Corpus/ClassiCC-PT}}, serves as the foundation for subsequent experimental analyses. Figure~\ref{fig:process} illustrates the overall process of creating ClassiCC-PT and will be discussed in further detail as follows.

\subsubsection{Retrieving Portuguese Pages}

Common Crawl periodically snapshots the web and offers metadata for each document in the snapshot, including language statistics computed using the CLD2 Naïve Bayesian classifier\footnote{\url{https://github.com/CLD2Owners/cld2}}. On average, Portuguese accounts for approximately 2\% of the pages in any given crawl. Using the provided metadata, we filter and select only the pages that include Portuguese in their language list. We then retrieve the corresponding WARC files for the specified crawl and extract the raw HTML content from each selected page.

\subsubsection{From HTML to Text}

The next step in creating the dataset involves extracting text from the HTML. Although Common Crawl provides a cleaned text version of the web pages through their WET files\footnote{\url{https://commoncrawl.org/blog/web-archiving-file-formats-explained}}, we opted to retrieve the raw HTML to investigate the impact of different extraction methods.

Accurately retrieving the main information from an HTML page is a complex task. Some datasets adopt a straightforward approach by simply removing markup and non-text elements from the HTML and then focusing on filtering high-quality web pages. More sophisticated methods have also been explored, such as rendering the visual representation of the web page and extracting text from it or parsing the HTML DOM tree and classifying each node as relevant content or noise~\citep{overwijk2022clueweb22}. These classifications can be performed heuristically, as seen in Trafilatura~\citep{trafilatura}, or through neural networks, as demonstrated in~\citep{neuscraper}.

We investigated the impact of extracting text using the naive approach of extracting all the text in the HTML and using Trafilatura. We observed that the naive approach generated significantly more tokens on average. Specifically, the naive method led to the extraction of approximately 100 billion tokens per crawl, while using Trafilatura resulted in an average of 45 billion tokens. 

After extracting the text from the HTML, our next step is to deduplicate our data since many studies have shown that duplicate data can negatively impact an LLM's performance~\citep{lee2021deduplicating,abbas2023semdedup}. We employed Minhash Deduplication on the documents from each crawl. We chose to perform only intra-crawl deduplication, following the findings of~\citet{fineweb}, which indicated that inter-crawl deduplication can decrease performance by retaining primarily high-entropy noise pages. On average, our intra-crawl deduplication process removed 40\% of the remaining web pages. Table~\ref{tab:dataset_docs_and_tokens} shows the final token count in ClassiCC-PT for each snapshot used.

In the result section, we will discuss the impact of each processing step. Our final dataset comprises 125 billion Portuguese tokens, comparable to other major open-source Portuguese corpora, such as mC4~\citep{mc4} (160 billion tokens), ClueWeb22-A~\citep{overwijk2022clueweb22}, which contains approximately 100 billion tokens, and Gigaverbo~\citep{correa2024tucano}, which contains 130 billion unique tokens.

\subsubsection{Filtering: Rule-Based Page Selection}

Training directly on web pages without applying quality filtering can often lead to suboptimal model performance~\citep{wenzek2019ccnet,gao2020pile,c4_intro}. To mitigate this issue, numerous studies have investigated rule-based methods to improve the quality of web page selection.

Rule-based selection involves categorizing a document by analyzing specific lexical attributes. These attributes may include the document's character count, the ratio of punctuation marks to characters, the number of stopwords present, and other relevant metrics.

We will examine the impacts of two sets of heuristic rules: those used in MassiveWeb~\citep{goupher_massiveweb} and those in C4~\citep{c4_intro, mc4}. We aim to understand the influence of these selection criteria and whether their effects are dataset-dependent. From the C4 rules, we specifically chose to apply only the subset that does not involve editing the document, which means that we excluded the following rules:

\begin{itemize}

\item Remove all lines not ending with a punctuation mark.

\item Deduplicate any three-sentence span in the dataset.

\item Remove any paragraphs containing ``privacy policy'', ``terms and conditions'' and similar phrases.

\end{itemize}

The complete set of rules for each rule set can be found in Appendix~\ref{sec:appendixA}. We observed that the MassiveWeb rules remove approximately 20\% of the documents in our dataset, while the C4 rules are more aggressive, resulting in a removal of around 43\% of documents. In the following sections, we discuss the impacts of these selections.

\subsubsection{Filtering: Neural Network-Based Selection}

The heuristic rules examined so far are effective, but predominantly focus on the lexical analysis of the document structure. However, when selecting data for pretraining a model, the type of content present is also crucial. FineWeb~\citep{fineweb} demonstrated impressive results by selecting documents with educational content from a large corpus. This selection process employs a linear regression model built on top of Snowflake-arctic-embed-m~\cite{snowflake_embed}. The linear classifier is trained to assess the educational value of a document based on synthetic annotations from Llama~3.

We performed our own synthetic annotation of Portuguese documents, randomly sampled from ClassiCC, labeling a total of 120 thousand documents based on their educational content. We used the same prompt as FineWeb, translated into Portuguese, and GPT-4o instead of Llama-3 to perform the annotation, a choice based on preliminary tests. We set aside 10k of these annotated samples as a test set, using the remaining 110k as a training set.

We tested the classifier developed by~\citet{fineweb} in our test set of Portuguese documents, and observed a significantly inferior performance in Portuguese compared to its results in English. This discrepancy is expected, as the classifier is a small model trained exclusively on English data. Consequently, this motivated us to develop our version of the classifier tailored to Portuguese.

We followed the the FineWeb formula, training a linear regressor on top of a model embeddings, we chose to build our model on top of BERTimbau~\citep{bertimbau} embeddings, since it yielded the best results in our preliminary results. We used our training set of 110k Portuguese annotated documents. Our final classifier, ClassiCC-PT-edu~\footnote{\url{https://huggingface.co/ClassiCC-Corpus/ClassiCC-PT-edu-classifier}}, was trained for 20 epochs using a cosine decay schedule with a 5\% warmup. We employed the AdamW optimizer~\citep{loshchilov2017decoupled} and learning rate of $3e^{-4}$.


We also developed two additional classifiers: one for STEM-related content, ClassiCC-PT-STEM~\footnote{\url{https://huggingface.co/ClassiCC-Corpus/ClassiCC-PT-stem-classifier}}, and another for identifying offensive content, ClassiCC-PT-toxic~\footnote{\url{https://huggingface.co/ClassiCC-Corpus/ClassiCC-PT-toxic-classifier}}. The performance of these classifiers will be discussed in more detail in the subsequent sections.

Our approach to creating the classifiers is similar to the one used in Gigaverbo.~\citep{correa2024tucano}, but our methodologies have key differences. While their classifier primarily assesses text quality in general\footnote{Tucano annotates documents with GPT-4o, asking it to score documents from 0 to 1, 'considering how reasonable, valuable, and informative this text is
for training a language model in Portuguese'.}, our classifiers focus on more specific objectives, such as measuring the presence of STEM-related content. We opted for these more niche objectives (STEM, educational, and toxic content) based on our preliminary tests, which indicated that broader objectives, such as evaluating overall quality of a document, can lead to inconsistencies in LLM annotation, given that the concept of ``text quality'' is inherently ambiguous without further specification.

Furthermore, we use a discrete scoring system when prompting GPT-4o to assess the educational content of a text on a scale from 0 to 5. A guideline is provided to assist in scoring; the prompt used is available in Appendix~\ref{sec:appendixA}, and further details about the annotation result can be found at Appendix~\ref{appendix:score_dist}. We then train our classifier to reproduce these scores. 

All documents in ClassiCC-PT have associated scores from each of our three classifiers. These scores provide an extra signal for future research, indicating their educational relevance, STEM focus, and potential toxicity.



\subsection{Model Training}

In this work, we explore the impacts of numerous steps in creating a large dataset for continued LLM training. To study these impacts, we chose to train a TinyLlama model trained in 1 trillion tokens in English\footnote{\url{https://huggingface.co/TinyLlama/TinyLlama-1.1B-intermediate-step-480k-1T}}. We selected this model for the following reasons:

\begin{itemize}

\item TinyLlama is a small model with 1.1B parameters, allowing us to perform multiple training runs with more than 100 billion tokens each.

\item It utilizes a recent model architecture in Llama2~\citep{touvron2023llama2}.

\item We have complete information on the type of data TinyLlama was trained on, which allows us to better discuss the results.

\item The TinyLlama project provided intermediary checkpoints during its development. We chose to start from the checkpoint trained on 1 trillion English tokens for our work. This checkpoint serves as a convenient comparison point, as our training dataset is approximately 10\% the size of the original training dataset.

\end{itemize}

We trained on TPUs v2-128, which was generously granted to us by Google's TRC program. We used the seqio and t5x~\citep{roberts2023t5x} frameworks to train and the lm-evaluation-harness framework~\citep{eval-harness} to perform our evaluations.

Our training runs started from the TinyLlama checkpoint trained by 1 trillion tokens in English; we use a learning rate of 1e-3 with the Adafactor optimizer~\citep{shazeer2018adafactor}, batch size of 256, and sequence length of 4096 tokens using packing.

\subsection{Evaluation}

A fundamental aspect of our study is to evaluate how well the trained LLMs perform on Portuguese tasks. We chose to use the Poeta benchmark~\citep{sabia}.

The Poeta benchmark consists of 14 diverse Portuguese tasks that cover a range of challenges and was first introduced alongside the Sabiá models~\citep{sabia}. A detailed list of these tasks and their respective types is provided in Table~\ref{tab:poeta_tasks}.

\begin{table*}[!htb]
\setlength{\tabcolsep}{1.2mm}
\centering
\caption{Tasks in the Poeta Benchmark.}
\label{tab:poeta_tasks}
\begin{tabular}{llcccc}
\toprule
\multirow{2}{*}{\textbf{Dataset}}       &    \multirow{2}{*}{\textbf{Type}}                            & \textbf{Preferred} & \textbf{Rand.} &    \multirow{2}{*}{\textbf{Transl.}} & \textbf{Num} \\ 
 &  & \textbf{Metric} & \textbf{Score} &  & \textbf{Few-Shot} \\
\midrule
AG News~\citep{agnews}        & Multiclass classification (4)  & Accuracy           & 25             & Yes & 12           \\
ASSIN 2 RTE~\citep{real2020assin}   & Binary classification          & F1                 & 50             & No  & 18           \\ 
ASSIN 2 STS~\citep{real2020assin}    & Regression                     & Pearson            & 0              & No  & 15           \\ 
BLUEX~\citep{almeida2023bluex}          & Multiple choice (4)            & Accuracy           & 25             & No  & 1            \\ 
BoolQ~\citep{clark-etal-2019-boolq}          & Binary classification          & Accuracy           & 50             & Yes & 4            \\
ENEM Challenge~\citep{silveira2018enem} & Multiple choice (5)            & Accuracy           & 20             & No  & 1            \\
ENEM 2022~\citep{enem2022}      & Multiple choice (5)            & Accuracy           & 20             & No  & 1            \\
FaQuAD~\citep{sayama2019faquad}         & Extractive QA                  & F1                 & 0              & No  & 4            \\ 
IMDB~\citep{imdb}          & Binary classification          & Accuracy           & 50             & Yes & 2            \\ 
MASSIVE~\citep{fitzgerald2022massive}        & Multiclass classification (18) & F1-macro           & 0.58           & Yes & 36           \\
MKQA~\citep{longpre2021mkqa}           & Extractive QA                  & F1                 & 0              & Yes & 40           \\
SST2~\citep{sst2}           & Binary classification          & Accuracy           & 50             & Yes & 34           \\
TweetSentBR~\citep{tweetsentbr}    & Multiclass classification (3)  & F1-macro           & 32.4           & No  & 30           \\
WSC~\citep{sakaguchi2021winogrande}            & Binary classification          & Accuracy           & 50             & Yes & 18           \\
\bottomrule
\end{tabular}
\end{table*}

We use the Normalized Preferred Metric (NPM), the same aggregation metric as in the Poeta benchmark~\citep{sabia}. NPM provides an overall performance summary by normalizing each task score based on its expected random and maximum score. This ensures that binary classification tasks, with a 50\% random baseline, does not disproportionately affect the average compared to a task with a baseline of 0\%. The NPM is calculated as follows:
\begin{equation*}
\text{NPM} = \frac{1}{N} \sum_{i=1}^N 100 \times \frac{\text{[preferred metric]}_i - \text{[random score]}_i}{\text{[max score]}_i - \text{[random score]}_i} \PeriodPunct
\end{equation*}

To illustrate the benefits of using NPM, consider a multiple-choice task with four alternatives, where random guessing yields an expected accuracy of 25\%, alongside an open-ended question answering task, where random performance is 0\%. If we simply averaged the raw accuracies, a model guessing randomly would score 12.5\% across the two tasks. In contrast, with NPM, both random baselines are normalized to 0, yielding an average NPM of 0 for a random model, which enhances interpretability. This normalization enables fairer aggregation across tasks with different baseline difficulties and supports combining various preferred metrics, such as accuracy, F1, or exact match, within a unified framework.

\section{Results} 

In this section, we describe the observed impacts of numerous aspects of data curation mentioned previously.

\subsection{Impacts of Data Processing}

\begin{figure}[!htb]
\centering
\includegraphics[width=0.97\columnwidth]{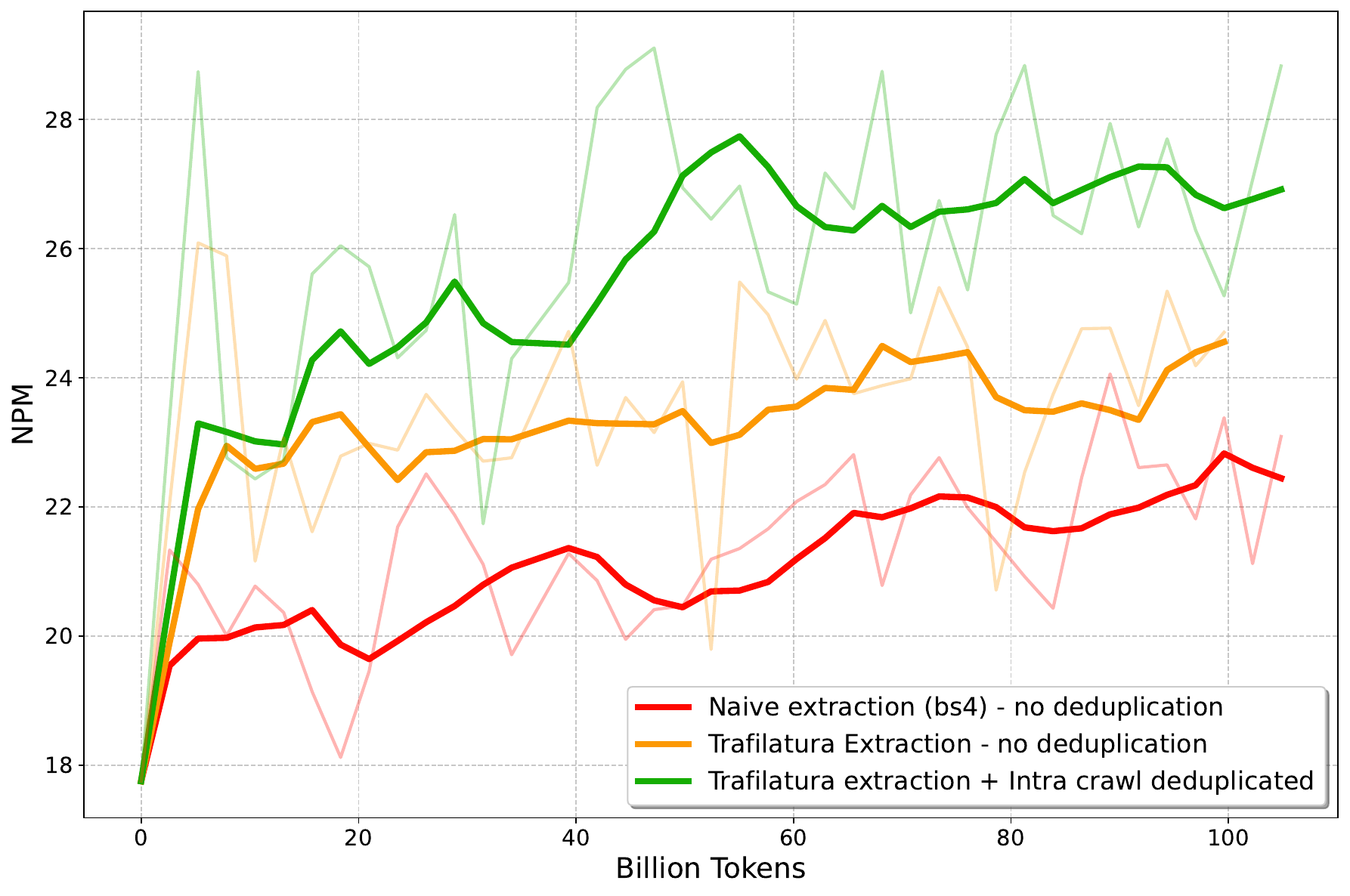}
\caption{NPM of three different runs. The red line shows a run using all the text available at the HTML, the orange line shows the a run using the text extracted using the Trafilatura library, and the green line shows a run using the text extracted with the Trafilatura library and deduplicated intra crawl.}
\label{fig:extraction_and_deduplication}
\end{figure}

Figure~\ref{fig:extraction_and_deduplication} presents the NPM in the Poeta benchmark for three different runs. Each run utilized data from the same three selected Portuguese subset crawls from CC, but with different preprocessing strategies applied:

\begin{itemize}

\item Run 1: All text content was extracted directly from the HTML using the BeautifulSoup library.

\item Run 2: Text was extracted from the HTML using the Trafilatura library, which applies content-cleaning rules.

\item Run 3: Text extraction was performed with Trafilatura, followed by an additional intra-crawl deduplication step using MinHash.

\end{itemize}

The results show a performance improvement when using Trafilatura for text extraction compared to extracting all text with BeautifulSoup. This improvement is expected because Trafilatura employs sophisticated heuristics to remove noisy or irrelevant content (such as headers, footers, and advertisements) from HTML pages~\citep{trafilatura}. By reducing the amount of non-informative or low-quality text, Trafilatura provides a more refined dataset, contributing to better model performance.

Another improvement was observed when applying intra-crawl MinHash deduplication to the text extracted by Trafilatura. This deduplication step is crucial for mitigating the negative effects of excessive duplicate content in training data. Consistent with findings from previous studies~\citep{lee2021deduplicating}, high levels of redundancy in training datasets can degrade model performance by introducing biases in the learning process or causing overfitting to repetitive patterns. In our case of continued pretraining in Portuguese, we observed the same detrimental impact of duplicate content.

\subsection{Impacts of Data Selection}

This section discusses how rule-based and neural network-based selection approaches impacted our experiments and the performance of our classifiers.

\subsubsection{Rule-Based Selection}

To assess the impact of rule-based data selection methods, we applied two distinct filtering rule sets in our dataset: the MassiveWeb~\citep{goupher_massiveweb} rules and our adapted C4 rules~\citep{c4_intro}. We then trained models using datasets generated by each of these rule sets and the original dataset without any filtering.

All models were trained following a consistent methodology, starting from the TinyLlama checkpoint, which had been pretrained on 1 trillion tokens. The initial dataset contained a total of 120 billion tokens, which was reduced to 93 billion unique tokens after applying the MassiveWeb filtering rules. Further applying the C4 rules reduced the dataset size to 78 billion unique tokens.

\begin{figure}[!htb]
\centering
\includegraphics[width=0.97\columnwidth]{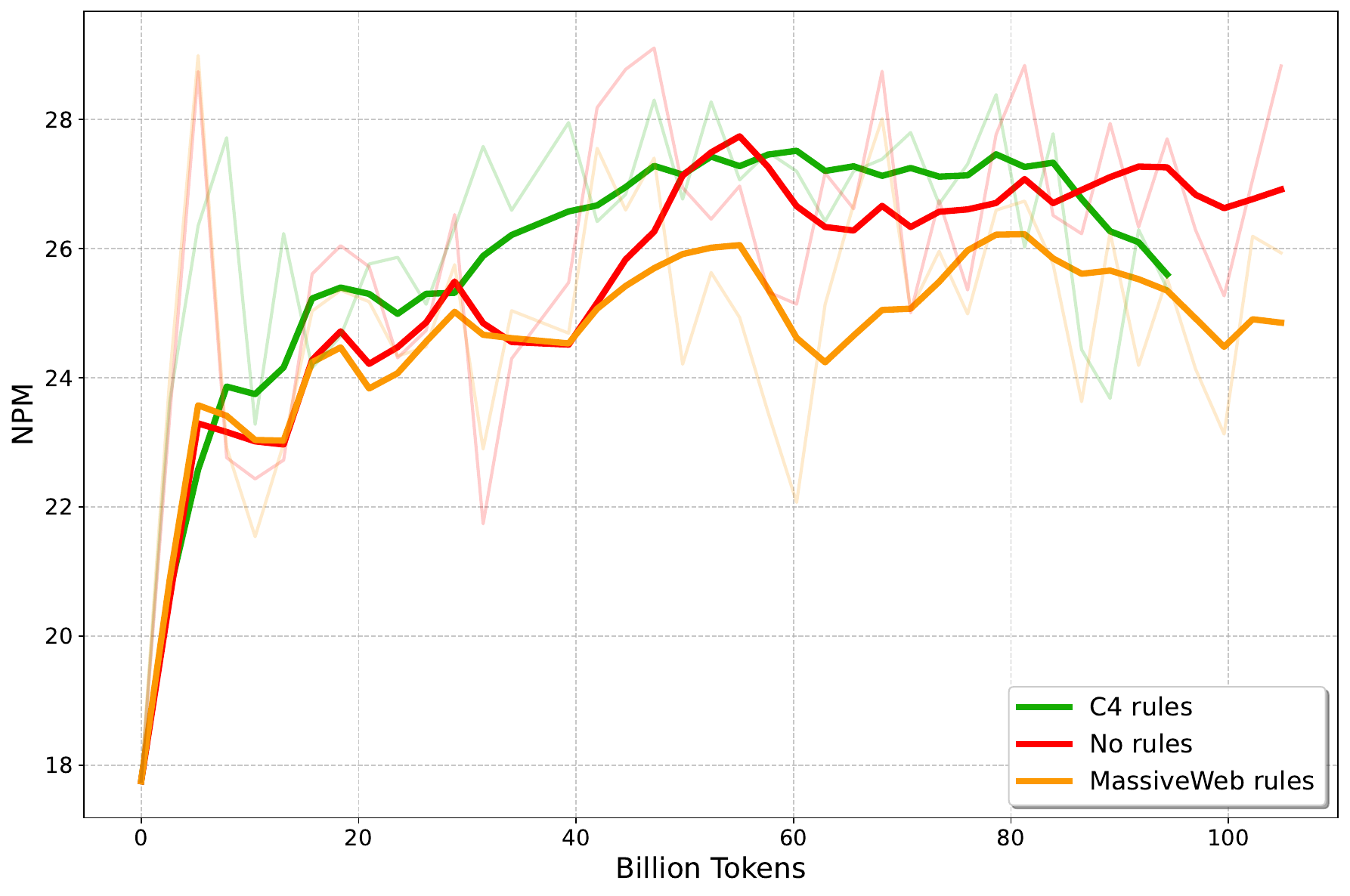}
\caption{NPM over three different runs, all using our dataset, but with different document selection-based rules applied.}
\label{fig:rules_impact}
\end{figure}

Figure~\ref{fig:rules_impact} presents the results of these experiments. The findings indicate that applying the C4 rules enhances model performance during training, up to the point where 80 billion tokens have been processed. Beyond this threshold, performance begins to decline, coinciding with the completion of a full epoch of the dataset filtered with the C4 rules. These results suggest that the C4 filtering rules are particularly beneficial in resource-constrained scenarios, allowing for effective training with fewer tokens than the total dataset size.

In contrast, datasets filtered using the MassiveWeb rules exhibited a slight decline in performance compared to the unfiltered baseline. This suggests that the MassiveWeb rules may have inadvertently removed documents containing useful or relevant information, thereby reducing the overall quality of the training data. This finding conflicts with previous studies~\citep{goupher_massiveweb,sabia}. However, we believe this behavior may be a consequence of our specific setup, which involves training a small model with no prior pretraining in the target language.


\subsubsection{ClassiCC Classifiers}

Starting with our classifier results, we evaluate their alignment with the scores assigned by GPT-4o on a separate test set. Since the classifiers are regressors and the GPT-4o scores are discrete, we round each document's predicted score to the nearest integer to compute the F1 score between the classifier's outputs and the GPT-4o scores.

Table~\ref{tab:classifier_edu} compares the original FineWeb-Edu classifier and ClassiCC-PT-Edu when scoring Portuguese documents. The FineWeb-Edu classifier does not perform well in Portuguese, which is expected since it was trained only in English. In practice, the classifier scores are typically used to distinguish ``good'' and ``bad'' documents by applying a threshold and retaining only documents with scores above this threshold. We also analyze this scenario in Table~\ref{tab:classifier_edu}, assuming a threshold of 3, the same as that used in FineWeb~\citep{fineweb}. Under this threshold, ClassiCC-PT-Edu achieves an F1 score of 0.77, which is competitive with the performance of the FineWeb-edu classifier in English.

\begin{table}[!htb]
\setlength{\tabcolsep}{2mm}
\centering
\caption{Results for educational content classification in Portuguese. The test set consists of 10k examples.}
\label{tab:classifier_edu}
\begin{tabular}{lccl}
\toprule
\textbf{Classifier Model} & \textbf{Precision} & \textbf{Recall} & \textbf{F1} \\
\midrule 
\multicolumn{4}{c}{Multiclass} \\
\midrule 
ClassiCC-PT-Edu & 0.46 & 0.44 & 0.43 \\
FineWeb-edu & 0.16 & 0.19 & 0.13 \\
\midrule 
\multicolumn{4}{c}{Binary (Threshold 3)} \\
\midrule
ClassiCC-PT-Edu & 0.77 & 0.77 & 0.77 \\
FineWeb-Edu & 0.45 & 0.50 & 0.48 \\
\bottomrule
\end{tabular}
\end{table}

As previously mentioned, we trained two additional classifiers for STEM content and for detecting toxic and offensive content. To achieve this, we followed the same process used for the educational classifier, annotating many documents using GPT-4o with the corresponding prompts. The F1 scores for the respective test sets are reported in Table~\ref{tab:all_classifiers}, showing that all classifiers achieve similar F1 scores after binarization.

\begin{table}[!htb]
\setlength{\tabcolsep}{1mm}
\centering
\caption{Performance for all ClassiCC-PT classifiers using a binary threshold of 3.}
\label{tab:all_classifiers}
\begin{tabular}{cccc}
\toprule 
\textbf{Classifier Model} & \textbf{F1} & \textbf{Test Size} & \textbf{Training Size} \\
\midrule 
ClassiCC-PT-edu & 0.77&10k & 110k \\
ClassiCC-PT-stem  & 0.76 & 12k & 100k \\ 
ClassiCC-PT-toxic  &  0.78 & 20k & 180k \\
\bottomrule
\end{tabular}
\end{table}

\subsubsection{Neural Network-Based Selection}

In our study, we developed classifiers to categorize document content, focusing specifically on STEM and education-related materials. Through this categorization, we found that only about 10\% of the documents in our dataset received a high education or STEM score. This relatively small proportion posed a challenge, limiting our ability to assess their isolated impact on the model's performance.

\begin{figure}[!htb]
\centering
\includegraphics[width=0.97\columnwidth]{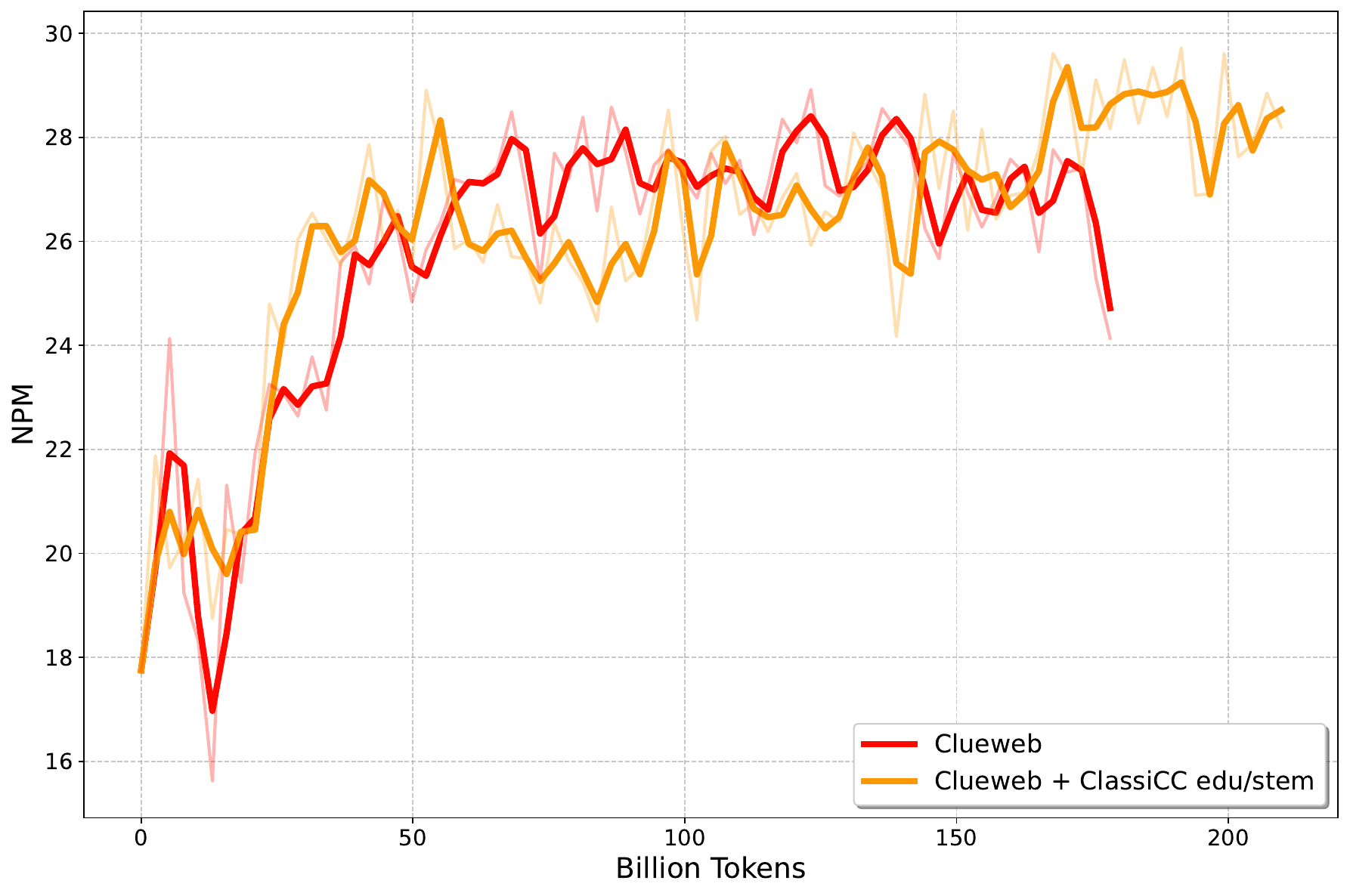}
\caption{NPM of two different runs. The orange line shows the performance using only the ClueWeb dataset, the orange line shows the performance when mixing the ClueWeb dataset to our selected high quality documents.}
\label{fig:classifiers_clueweb}
\end{figure}

We conducted an experiment using a mixture of datasets to assess whether the educational and STEM-focused documents contributed valuable information to the model. The baseline dataset for this comparison was a subset of the ClueWeb dataset, containing a total of 80 billion tokens. We then created a mixed dataset by combining this ClueWeb subset with the educational and STEM documents from ClassiCC-PT, increasing the total token count to approximately 100 billion. Using these datasets, we trained two separate models -- one on the ClueWeb subset and the other on the mixed dataset -- over two epochs. The results of this comparison are presented in Figure~\ref{fig:classifiers_clueweb}.

In the initial stages, up to 60 billion tokens, the model trained on the mixed dataset outperformed the ClueWeb-only model. However, between 60 billion and 150 billion tokens, the ClueWeb-only model exhibited better performance. Towards the end of the second epoch, the model trained on the mixed dataset began to significantly surpass the ClueWeb-only model, ultimately achieving a peak gain of approximately 1.5 NPM points over the baseline model.

These results highlight the potential value of incorporating targeted educational and STEM content into training datasets, even when such content constitutes a relatively small fraction of the overall dataset.


\subsection{Comparison with Other Datasets}

\begin{figure}[!htb]
\centering
\includegraphics[width=0.97\columnwidth]{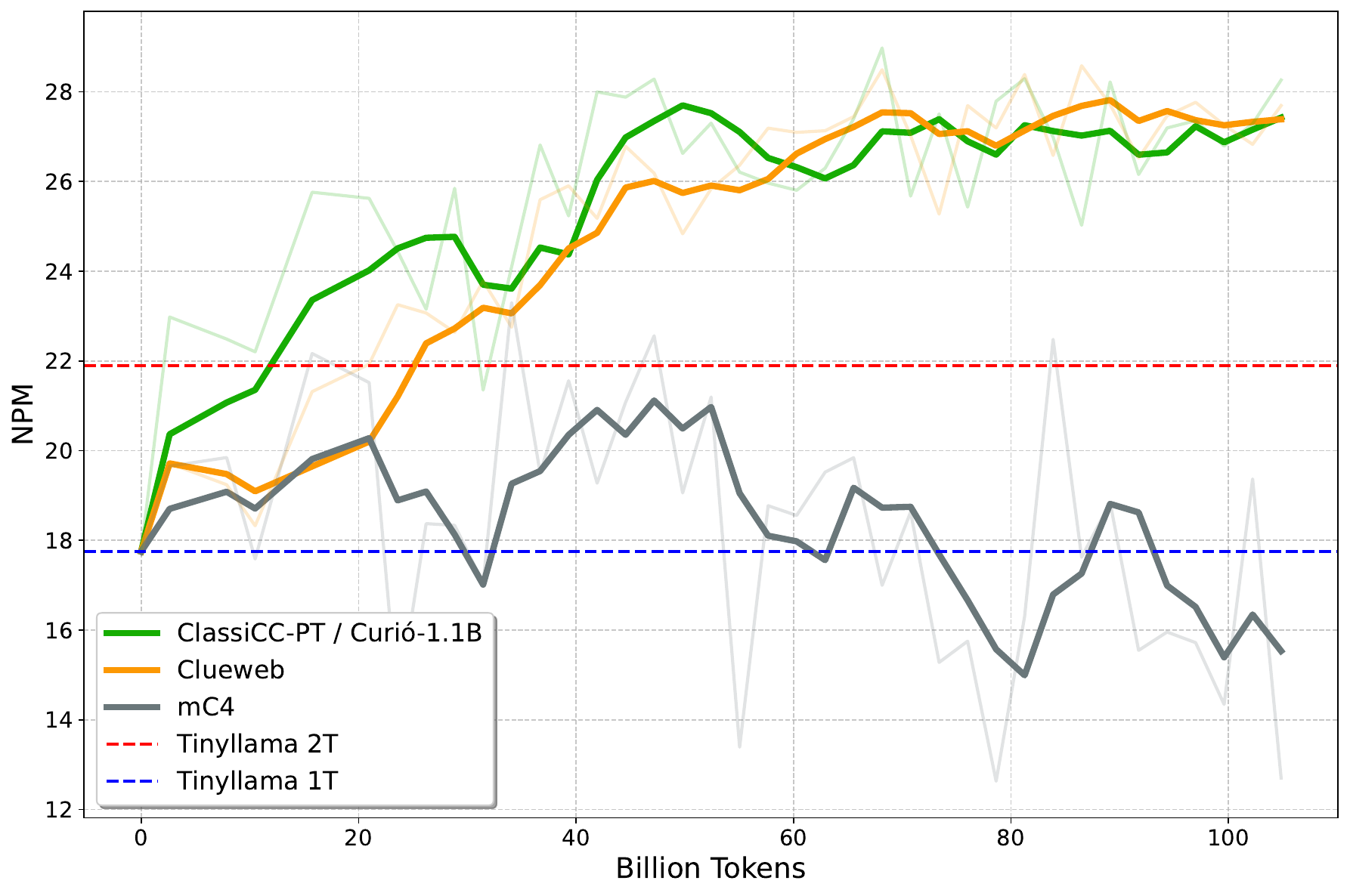}
\caption{NPM of three different runs using different datasets. The orange line shows the performance using the ClueWeb dataset, the gray line shows the performance using mC4, and the green line shows the performance using our dataset. The dotted lines indicate the base performance of TinyLlama trained in 1T and 2T tokens.}
\label{fig:all_datasets}
\end{figure}

Finally, we compare our dataset with two other Portuguese corpora of similar size, emphasizing the differences in their construction processes and performance outcomes. Figure~\ref{fig:all_datasets} illustrates the NPM performance across mC4, ClueWeb, and ClassiCC-PT. We also name the model trained on ClassiCC-PT Curi\'o 1.1B~\footnote{\url{https://huggingface.co/ClassiCC-Corpus/Curio-1.1b}}.

The results indicate that despite the relatively straightforward preprocessing steps used in constructing our dataset, it achieves a performance comparable to that of the ClueWeb dataset, which employs an industry-grade processing pipeline~\citep{overwijk2022clueweb22}.

ClassiCC-PT and ClueWeb significantly outperform the mC4 dataset, demonstrating that well-curated preprocessing strategies can yield high-quality datasets even without the resources and scale of commercial-level pipelines. This finding underscores the potential for researchers to create competitive datasets using more accessible techniques, thereby democratizing advancements in Portuguese-language NLP.

\subsection{Training from Scratch x Continued Pretraining}

To quantify the impact of adopting the continued pretraining paradigm, we conducted an experiment using ClassiCC-PT to train a model with the same architecture as TinyLlama but starting the weights from scratch.

Following the recommendations of PALM~\citep{chowdhery2023palm}, we employed an Adafactor learning rate of $10^{-2}$ for the first 10k steps, which was then decayed at a rate of $1/\sqrt{k}$, where $k$ is the step number. We also used a batch size of 256, consistent with the continued pretraining experiment. This experiment ran for 2 epochs on the ClassiCC-PT dataset.

\begin{figure}[!htb]
\centering
\includegraphics[width=0.97\columnwidth]{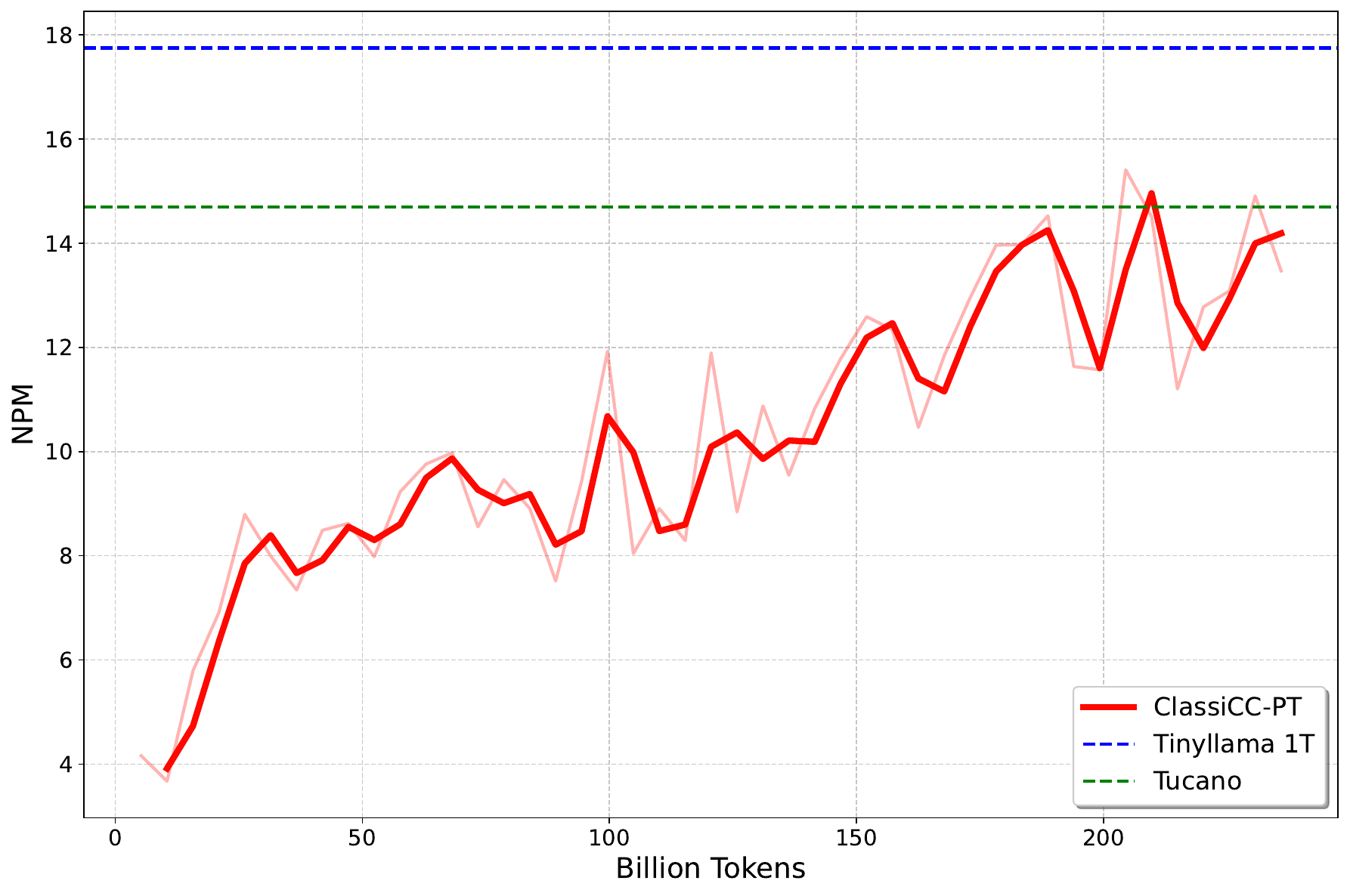}
\caption{NPM during training of model from scratch using ClassiCC-PT.}
\label{fig:from_scratch_training}
\end{figure}

Figure~\ref{fig:from_scratch_training} illustrates the NPM during our from-scratch training. The gains are generally increasing but show high variance. Notably, around the 200 billion tokens mark, approaching the end of the two epochs, some checkpoints demonstrate performance comparable to Tucano 1.1B, a recent Portuguese LLM trained from scratch on 250 billion tokens using the Gigaverbo dataset.

Table~\ref{tab:performance_per_train_type} presents the performance of different training regimens regarding the achieved NPM on the Poeta benchmark. We observe that models trained from scratch exclusively in Portuguese, even with extended training, struggle to achieve high performance.

\begin{table*}[!htb]
\setlength{\tabcolsep}{4mm}
\centering
\caption{Comparison of performance of different training regimens, all models from scratch trained only in Portuguese achieved similar NPMs. Starting from TinyLlama yielded great benefits, achieving better results than models trained in double the amount of Portuguese tokens. All models listed in the table use Llama architecture.}
\label{tab:performance_per_train_type}
\begin{tabular}{lcccc}
\toprule
\multirow{ 2}{*}{\textbf{Model}}                & \textbf{Starting} & \textbf{Tokens}  & \textbf{Poeta}   & \textbf{Training} \\
                & \textbf{Weights} & \textbf{Trained}  & \textbf{(NPM)}   & \textbf{Dataset} \\
\midrule
Curió 1.1B    & TinyLlama 1T     & 1T EN + 120B PT & \textbf{27.1} & ClassiCC-PT (1 epoch) \\
Ours ( from scratch) & Scratch          & 120B PT         & 11.7          & ClassiCC-PT (1 epoch) \\
Ours ( from scratch) & Scratch          & 240B PT         & 14.9          & ClassiCC-PT (2 epoch) \\
Tucano 1.1B          & Scratch          & 250B PT         & 14.9          & Gigaverbo (1.5 epochs)  \\
TinyLlama 1T         & Scratch          & 1T EN           & 17.4          & SlimPajama - 1T tokens  \\
TinyLlama 2T         & Scratch          & 2T EN           & 20.9          & SlimPajama - 2T tokens  \\
\bottomrule
\end{tabular}
\end{table*}

Our model, trained from scratch on 120 billion Portuguese tokens, achieves an NPM of 11.7. Doubling the training size by performing an additional epoch on the ClassiCC-PT dataset yields an extra 3.2 NPM points, bringing the total to 14.9. This score is comparable to that achieved by Tucano 1.1B~\citep{correa2024tucano}, which is expected, as both models were trained on a similar number of Portuguese tokens, although using different datasets. It is worth noting that these results are somewhat competitive with TinyLlama 1T, which was trained on four times more tokens in English.

Regarding continued pretraining, Curió 1.1B, which starts from TinyLlama 1T, displays a significant advantage and achieves an impressive NPM of 27.1. This result suggests that leveraging existing models, even those trained primarily on English data, provides a stronger foundation for continued training in Portuguese than starting from scratch, even when using twice as many in-language tokens.

\section{Conclusions} 

This study presented the creation, processing, and evaluation of a large-scale Portuguese dataset for training language models. Starting from the CommonCrawl corpus, we showed that a simple method can produce a dataset that competes with industry-standard corpora.

Through continued pretraining experiments with the TinyLlama model, we showed that well-curated data plays a key role in improving performance on Portuguese benchmarks such as Poeta. Our dataset performed on par with the Portuguese portion of ClueWeb-22, which was built using Bing infrastructure, and outperformed other common datasets such as mC4. This highlights the value of our preprocessing approach.

Lastly, we demonstrated the importance of localized data. Both Curió-1.1B and TinyLlama 2T started from the same base model (TinyLlama 1T), but followed different training paths: Curió-1.1B was further pretrained on 100 billion Portuguese tokens, while TinyLlama 2T continued training on 1 trillion English tokens. Curió-1.1B achieved significantly better results on Portuguese tasks, highlighting the value of language-specific pretraining.

\section*{Declarations}

\begin{acknowledgements}
The authors would like to thank the Google Cloud TRC program for the generous TPU grant that made this research possible. They are also grateful to Maritaca AI for providing us with the necessary computation for most of our studies.
\end{acknowledgements}

\begin{funding}
The computational resources used in this research were funded by the Google TRC program (TPUs) and Maritaca AI (general computation and GPUs).
\end{funding}

\begin{contributions}
TSA contributed to the study's conception, data curation, code implementation, formal analysis, experiment execution, result evaluation, and manuscript preparation. RN and HP contributed by editing and reviewing the manuscript, securing funding, and supervising the research. All authors have read and approved the final version of the manuscript.
\end{contributions}








\bibliographystyle{apalike-sol}  
\bibliography{references.bib}

\begin{thebibliography}{}

\bibitem[Abbas {\em et~al}., 2023]{abbas2023semdedup}
Abbas, A., Tirumala, K., Simig, D., Ganguli, S., and Morcos, A.~S. (2023).
\newblock {SemDeDup: Data-Efficient Learning at Web-Scale through Semantic Deduplication}.
\newblock {\em arXiv preprint arXiv:2303.09540}, pages 1--34.

\bibitem[Abonizio {\em et~al}., 2024]{sabia3}
Abonizio, H., Almeida, T.~S., Laitz, T., {Malaquias Junior}, R., Bon{\'a}s, G.~K., Nogueira, R., and Pires, R. (2024).
\newblock {Sabi\'a Technical Report}.
\newblock {\em arXiv preprint arXiv:2410.12049}, pages 1--16.

\bibitem[Achiam {\em et~al}., 2023]{achiam2023gpt4_technical}
Achiam, J., Adler, S., Agarwal, S., Ahmad, L., Akkaya, I., Aleman, F.~L., Almeida, D., Altenschmidt, J., Altman, S., Anadkat, S., {\em et~al}. (2023).
\newblock {GPT-4 Technical Report}.
\newblock {\em arXiv preprint arXiv:2303.08774}.

\bibitem[Almeida {\em et~al}., 2024]{sabia2}
Almeida, T.~S., Abonizio, H., Nogueira, R., and Pires, R. (2024).
\newblock {Sabi\'a-2: A New Generation of Portuguese Large Language Models}.
\newblock {\em arXiv preprint arXiv:2403.09887}, pages 1--21.

\bibitem[Almeida {\em et~al}., 2025]{almeida2025tiebe}
Almeida, T.~S., Bon{\'a}s, G.~K., Santos, J. G.~A., Abonizio, H., and Nogueira, R. (2025).
\newblock {TiEBe: A Benchmark for Assessing the Current Knowledge of Large Language Models}.
\newblock {\em arXiv preprint arXiv:2501.07482}, pages 1--15.

\bibitem[Almeida {\em et~al}., 2023]{almeida2023bluex}
Almeida, T.~S., Laitz, T., Bon{\'a}s, G.~K., and Nogueira, R. (2023).
\newblock {BLUEX: A Benchmark Based on Brazilian Leading Universities Entrance Exams}.
\newblock In {\em Brazilian Conference on Intelligent Systems}, pages 337--347. Springer.

\bibitem[Azerbayev {\em et~al}., 2024]{azerbayev2024llemma}
Azerbayev, Z., Schoelkopf, H., Paster, K., Santos, M.~D., McAleer, S.~M., Jiang, A.~Q., Deng, J., Biderman, S., and Welleck, S. (2024).
\newblock {Llemma: An Open Language Model for Mathematics}.
\newblock In {\em The Twelfth International Conference on Learning Representations}, pages 1--28.

\bibitem[Bai {\em et~al}., 2023]{qwen}
Bai, J., Bai, S., Chu, Y., Cui, Z., Dang, K., Deng, X., Fan, Y., Ge, W., Han, Y., Huang, F., {\em et~al}. (2023).
\newblock {Qwen Technical Report}.
\newblock {\em arXiv preprint arXiv:2309.16609}, pages 1--59.

\bibitem[Bai {\em et~al}., 2022]{bai2022constitutional}
Bai, Y., Kadavath, S., Kundu, S., Askell, A., Kernion, J., Jones, A., Chen, A., Goldie, A., Mirhoseini, A., and McKinnon, C. (2022).
\newblock {Constitutional AI: Harmlessness from AI feedback}.
\newblock {\em arXiv preprint arXiv:2212.08073}, pages 1--34.

\bibitem[Barbaresi, 2021]{trafilatura}
Barbaresi, A. (2021).
\newblock {Trafilatura: A Web Scraping Library and Command-Line Tool for Text Discovery and Extraction}.
\newblock In {\em 59th Annual Meeting of the Association for Computational Linguistics and the 11th International Joint Conference on Natural Language Processing: System Demonstrations}, pages 122--131.

\bibitem[Brown {\em et~al}., 2020]{gpt3_base}
Brown, T., Mann, B., Ryder, N., Subbiah, M., Kaplan, J.~D., Dhariwal, P., Neelakantan, A., Shyam, P., Sastry, G., and Askell, A. (2020).
\newblock {Language Models are Few-Shot Learners}.
\newblock {\em Advances in Neural Information Processing Systems}, 33:1877--1901.

\bibitem[Brum and Volpe~Nunes, 2018]{tweetsentbr}
Brum, H. and Volpe~Nunes, M. d.~G. (2018).
\newblock {Building a Sentiment Corpus of Tweets in Brazilian Portuguese}.
\newblock In {\em Eleventh International Conference on Language Resources and Evaluation}, pages 1--5, Miyazaki, Japan. European Language Resources Association (ELRA).

\bibitem[Carlini {\em et~al}., 2021]{mixture_datasets_1}
Carlini, N., Tramer, F., Wallace, E., Jagielski, M., Herbert-Voss, A., Lee, K., Roberts, A., Brown, T., Song, D., and Erlingsson, U. (2021).
\newblock {Extracting Training Data from Large Language Models}.
\newblock In {\em 30th USENIX Security Symposium (USENIX Security 21)}, pages 2633--2650.

\bibitem[Chen {\em et~al}., 2023]{chen2023meditron}
Chen, Z., Cano, A.~H., Romanou, A., Bonnet, A., Matoba, K., Salvi, F., Pagliardini, M., Fan, S., K{\"o}pf, A., Mohtashami, A., {\em et~al}. (2023).
\newblock {Meditron-70B: Scaling Medical Pretraining for Large Language Models}.
\newblock {\em arXiv preprint arXiv:2311.16079}, pages 1--38.

\bibitem[Chowdhery {\em et~al}., 2023]{chowdhery2023palm}
Chowdhery, A., Narang, S., Devlin, J., Bosma, M., Mishra, G., Roberts, A., Barham, P., Chung, H.~W., Sutton, C., Gehrmann, S., {\em et~al}. (2023).
\newblock {PaLM: Scaling Language Modeling with Pathways}.
\newblock {\em Journal of Machine Learning Research}, 24(240):1--113.

\bibitem[Clark {\em et~al}., 2019]{clark-etal-2019-boolq}
Clark, C., Lee, K., Chang, M.-W., Kwiatkowski, T., Collins, M., and Toutanova, K. (2019).
\newblock {BoolQ: Exploring the Surprising Difficulty of Natural Yes/No Questions}.
\newblock In {\em Conference of the North American Chapter of the Association for Computational Linguistics: Human Language Technologies}, volume~1, pages 2924--2936, Minneapolis, Minnesota. Association for Computational Linguistics.

\bibitem[Colombo {\em et~al}., 2024]{saulm}
Colombo, P., Pires, T.~P., Boudiaf, M., Culver, D., Melo, R., Corro, C., Martins, A.~F., Esposito, F., Raposo, V.~L., Morgado, S., {\em et~al}. (2024).
\newblock {SaulLM-7B: A Pioneering Large Language Model for Law}.
\newblock {\em arXiv preprint arXiv:2403.03883}, pages 1--13.

\bibitem[Corr{\^e}a {\em et~al}., 2024]{correa2024tucano}
Corr{\^e}a, N.~K., Sen, A., Falk, S., and Fatimah, S. (2024).
\newblock {Tucano: Advancing Neural Text Generation for Portuguese}.
\newblock {\em arXiv preprint arXiv:2411.07854}, pages 1--34.

\bibitem[Dodge {\em et~al}., 2021]{c4Documenting}
Dodge, J., Sap, M., Marasovi{\'c}, A., Agnew, W., Ilharco, G., Groeneveld, D., Mitchell, M., and Gardner, M. (2021).
\newblock {Documenting Large Webtext Corpora: A Case Study on the Colossal Clean Crawled Corpus}.
\newblock In {\em Conference on Empirical Methods in Natural Language Processing}, pages 1286--1305.

\bibitem[Dubey {\em et~al}., 2024]{dubey2024llama3}
Dubey, A., Jauhri, A., Pandey, A., Kadian, A., Al-Dahle, A., Letman, A., Mathur, A., Schelten, A., Yang, A., and Fan, A. (2024).
\newblock {The Llama 3 Herd of Models}.
\newblock {\em arXiv preprint arXiv:2407.21783}, pages 1--92.

\bibitem[Endr{\'e}dy and Nov{\'a}k, 2013]{justext}
Endr{\'e}dy, I. and Nov{\'a}k, A. (2013).
\newblock {More Effective Boilerplate Removal-the Goldminer Algorithm}.
\newblock {\em Polibits}, 48:79--83.

\bibitem[FitzGerald {\em et~al}., 2022]{fitzgerald2022massive}
FitzGerald, J., Hench, C., Peris, C., Mackie, S., Rottmann, K., Sanchez, A., Nash, A., Urbach, L., Kakarala, V., and Singh, R. (2022).
\newblock {MASSIVE: A 1M-Example Multilingual Natural Language Understanding Dataset with 51 Typologically-diverse Languages}.
\newblock {\em arXiv preprint arXiv:2204.08582}, pages 1--24.

\bibitem[Gao {\em et~al}., 2020]{gao2020pile}
Gao, L., Biderman, S., Black, S., Golding, L., Hoppe, T., Foster, C., Phang, J., He, H., Thite, A., and Nabeshima, N. (2020).
\newblock {The Pile: An 800GB Dataset of Diverse Text for Language Modeling}.
\newblock {\em arXiv preprint arXiv:2101.00027}, pages 1--39.

\bibitem[Gao {\em et~al}., 2024]{eval-harness}
Gao, L., Tow, J., Abbasi, B., Biderman, S., Black, S., DiPofi, A., Foster, C., Golding, L., Hsu, J., Le~Noac'h, A., Li, H., McDonell, K., Muennighoff, N., Ociepa, C., Phang, J., Reynolds, L., Schoelkopf, H., Skowron, A., Sutawika, L., Tang, E., Thite, A., Wang, B., Wang, K., and Zou, A. (2024).
\newblock {A Framework for Few-Shot Language Model Evaluation}.

\bibitem[Gogoulou {\em et~al}., 2024]{gogoulou2024languageshift}
Gogoulou, E., Lesort, T., Boman, M., and Nivre, J. (2024).
\newblock {Continual Learning under Language Shift}.
\newblock In {\em International Conference on Text, Speech, and Dialogue}, pages 71--84. Springer.

\bibitem[Heafield, 2011]{heafield2011kenlm}
Heafield, K. (2011).
\newblock {KenLM: Faster and Smaller Language Model Queries}.
\newblock In {\em Sixth Workshop on Statistical Machine Translation}, pages 187--197.

\bibitem[Kreutzer {\em et~al}., 2022]{kreutzer2022quality}
Kreutzer, J., Caswell, I., Wang, L., Wahab, A., van Esch, D., Ulzii-Orshikh, N., Tapo, A., Subramani, N., Sokolov, A., and Sikasote, C. (2022).
\newblock {Quality at a Glance: An Audit of Web-Crawled Multilingual Datasets}.
\newblock {\em Transactions of the Association for Computational Linguistics}, 10:50--72.

\bibitem[Kudugunta {\em et~al}., 2024]{madlad}
Kudugunta, S., Caswell, I., Zhang, B., Garcia, X., Xin, D., Kusupati, A., Stella, R., Bapna, A., and Firat, O. (2024).
\newblock {MADLAD-400: A Multilingual and Document-Level Large Audited Dataset}.
\newblock {\em Advances in Neural Information Processing Systems}, 36:1--60.

\bibitem[Labrak {\em et~al}., 2024]{biomistral}
Labrak, Y., Bazoge, A., Morin, E., Gourraud, P.-A., Rouvier, M., and Dufour, R. (2024).
\newblock {BioMistral: A Collection of Open-Source Pretrained Large Language Models for Medical Domains}.
\newblock {\em arXiv preprint arXiv:2402.10373}, pages 1--17.

\bibitem[Le~Scao {\em et~al}., 2023]{bloom}
Le~Scao, T., Fan, A., Akiki, C., Pavlick, E., Ili{\'c}, S., Hesslow, D., Castagn{\'e}, R., Luccioni, A.~S., Yvon, F., Gall{\'e}, M., {\em et~al}. (2023).
\newblock {BLOOM: A 176b-Parameter Open-Access Multilingual Language Model}.
\newblock {\em arXiv 2211.05100}, pages 1--74.

\bibitem[Lee {\em et~al}., 2021]{lee2021deduplicating}
Lee, K., Ippolito, D., Nystrom, A., Zhang, C., Eck, D., Callison-Burch, C., and Carlini, N. (2021).
\newblock {Deduplicating Training Data Makes Language Models Better}.
\newblock {\em arXiv preprint arXiv:2107.06499}, pages 1--22.

\bibitem[Lewkowycz {\em et~al}., 2022]{lewkowycz2022solving}
Lewkowycz, A., Andreassen, A.~J., Dohan, D., Dyer, E., Michalewski, H., Ramasesh, V.~V., Slone, A., Anil, C., Schlag, I., Gutman-Solo, T., Wu, Y., Neyshabur, B., Gur-Ari, G., and Misra, V. (2022).
\newblock {Solving Quantitative Reasoning Problems with Language Models}.
\newblock In Oh, A.~H., Agarwal, A., Belgrave, D., and Cho, K., editors, {\em Advances in Neural Information Processing Systems}, pages 1--54.

\bibitem[Liu {\em et~al}., 2024]{liu2024datasets}
Liu, Y., Cao, J., Liu, C., Ding, K., and Jin, L. (2024).
\newblock {Datasets for Large Language Models: A Comprehensive Survey}.
\newblock {\em arXiv preprint arXiv:2402.18041}, pages 1--181.

\bibitem[Longpre {\em et~al}., 2021]{longpre2021mkqa}
Longpre, S., Lu, Y., and Daiber, J. (2021).
\newblock {MKQA: A Linguistically Diverse Benchmark for Multilingual Open Domain Question Answering}.
\newblock {\em Transactions of the Association for Computational Linguistics}, 9:1389--1406.

\bibitem[Longpre {\em et~al}., 2024]{longpre2024bridgingprovenancegap}
Longpre, S., Singh, N., Cherep, M., Tiwary, K., Materzynska, J., Brannon, W., Mahari, R., Dey, M., Hamdy, M., and Saxena, N. (2024).
\newblock {Bridging the Data Provenance Gap Across Text, Speech and Video}.
\newblock {\em arXiv preprint arXiv:2412.17847}, pages 1--70.

\bibitem[Lopes {\em et~al}., 2024]{gloria}
Lopes, R., Magalhães, J., and Semedo, D. (2024).
\newblock {Gl\'orIA: A Generative and Open Large Language Model for Portuguese}.
\newblock {\em arXiv 2402.12969}.
\newblock \url{https://arxiv.org/abs/2402.12969}.

\bibitem[Loshchilov, 2017]{loshchilov2017decoupled}
Loshchilov, I. (2017).
\newblock {Decoupled Weight Decay Regularization}.
\newblock {\em arXiv preprint arXiv:1711.05101}, pages 1--19.

\bibitem[Maas {\em et~al}., 2011]{imdb}
Maas, A., Daly, R.~E., Pham, P.~T., Huang, D., Ng, A.~Y., and Potts, C. (2011).
\newblock {Learning Word Vectors for Sentiment Analysis}.
\newblock In {\em 49th Annual Meeting of the Association for Computational Linguistics: Human Language Technologies}, pages 142--150.

\bibitem[Marion {\em et~al}., 2023]{marion2023less}
Marion, M., {\"U}st{\"u}n, A., Pozzobon, L., Wang, A., Fadaee, M., and Hooker, S. (2023).
\newblock {When Less is More: Investigating Data Pruning for Pretraining LLMs at Scale}.
\newblock {\em arXiv preprint arXiv:2309.04564}, pages 1--25.

\bibitem[Martins {\em et~al}., 2024]{eurollm}
Martins, P.~H., Fernandes, P., Alves, J., Guerreiro, N.~M., Rei, R., Alves, D.~M., Pombal, J., Farajian, A., Faysse, M., Klimaszewski, M., {\em et~al}. (2024).
\newblock {EuroLLM: Multilingual language Models for Europe}.
\newblock {\em arXiv preprint arXiv:2409.16235}, pages 1--12.

\bibitem[Merrick, 2024]{snowflake_embed}
Merrick, L. (2024).
\newblock Embedding and clustering your data can improve contrastive pretraining.
\newblock {\em arXiv preprint arXiv:2407.18887}.

\bibitem[Moayeri {\em et~al}., 2024]{moayeri2024worldbench}
Moayeri, M., Tabassi, E., and Feizi, S. (2024).
\newblock Worldbench: Quantifying geographic disparities in llm factual recall.
\newblock In {\em ACM Conference on Fairness, Accountability, and Transparency}, pages 1211--1228.

\bibitem[Nguyen {\em et~al}., 2024]{culturax}
Nguyen, T., Van~Nguyen, C., Lai, V.~D., Man, H., Ngo, N.~T., Dernoncourt, F., Rossi, R.~A., and Nguyen, T.~H. (2024).
\newblock {CulturaX: A Cleaned, Enormous, and Multilingual Dataset for Large Language Models in 167 Languages}.
\newblock In {\em Joint International Conference on Computational Linguistics, Language Resources and Evaluation (LREC-COLING 2024)}, pages 4226--4237.

\bibitem[Nguyen {\em et~al}., 2023]{seallm}
Nguyen, X.-P., Zhang, W., Li, X., Aljunied, M., Hu, Z., Shen, C., Chia, Y.~K., Li, X., Wang, J., Tan, Q., {\em et~al}. (2023).
\newblock {SeaLLMs: Large Language Models for Southeast Asia}.
\newblock {\em arXiv preprint arXiv:2312.00738}, pages 1--11.

\bibitem[Nunes {\em et~al}., 2023]{enem2022}
Nunes, D., Primi, R., Pires, R., Lotufo, R., and Nogueira, R. (2023).
\newblock {Evaluating GPT-3.5 and GPT-4 Models on Brazilian University Admission Exams}.
\newblock {\em arXiv preprint arXiv:2303.17003}.

\bibitem[Overwijk {\em et~al}., 2022]{overwijk2022clueweb22}
Overwijk, A., Xiong, C., and Callan, J. (2022).
\newblock {ClueWeb22: 10 Billion Web Documents with Rich Information}.
\newblock In {\em 45th International ACM SIGIR Conference on Research and Development in Information Retrieval}, pages 3360--3362.

\bibitem[Paul {\em et~al}., 2021]{pruning_2}
Paul, M., Ganguli, S., and Dziugaite, G.~K. (2021).
\newblock {Deep Learning on a Data Diet: Finding Important Examples Early in Training}.
\newblock {\em Advances in Neural Information Processing Systems}, 34:20596--20607.

\bibitem[Penedo {\em et~al}., 2024a]{fineweb}
Penedo, G., Kydl{\'\i}{\v{c}}ek, H., Lozhkov, A., Mitchell, M., Raffel, C., Von~Werra, L., Wolf, T., {\em et~al}. (2024a).
\newblock {The FineWeb Datasets: Decanting the Web for the Finest Text Data at Scale}.
\newblock {\em arXiv preprint arXiv:2406.17557}, pages 1--38.

\bibitem[Penedo {\em et~al}., 2024b]{penedo2024fineweb-2}
Penedo, G., Kydlíček, H., Sabolčec, V., Messmer, B., Foroutan, N., Jaggi, M., von Werra, L., and Wolf, T. (2024b).
\newblock {FineWeb2: A Sparkling Update with 1000s of Languages}.
\newblock \url{https://huggingface.co/datasets/HuggingFaceFW/fineweb-2}.

\bibitem[Penedo {\em et~al}., 2023]{penedo2023refinedweb}
Penedo, G., Malartic, Q., Hesslow, D., Cojocaru, R., Cappelli, A., Alobeidli, H., Pannier, B., Almazrouei, E., and Launay, J. (2023).
\newblock {The RefinedWeb Dataset for Falcon LLM: Outperforming Curated Corpora with Web Data, and Web Data Only}.
\newblock {\em arXiv preprint arXiv:2306.01116}, pages 1--32.

\bibitem[Pipatanakul {\em et~al}., 2023]{typhoon}
Pipatanakul, K., Jirabovonvisut, P., Manakul, P., Sripaisarnmongkol, S., Patomwong, R., Chokchainant, P., and Tharnpipitchai, K. (2023).
\newblock {Typhoon: Thai Large Language Models}.
\newblock {\em arXiv preprint arXiv:2312.13951}, pages 1--12.

\bibitem[Pires {\em et~al}., 2023]{sabia}
Pires, R., Abonizio, H., Almeida, T.~S., and Nogueira, R. (2023).
\newblock {Sabi\'a: Portuguese Large Language Models}.
\newblock In {\em Brazilian Conference on Intelligent Systems}, pages 226--240. Springer.

\bibitem[Rae {\em et~al}., 2021]{goupher_massiveweb}
Rae, J.~W., Borgeaud, S., Cai, T., Millican, K., Hoffmann, J., Song, F., Aslanides, J., Henderson, S., Ring, R., and Young, S. (2021).
\newblock {Scaling Language Models: Methods, Analysis \& Insights from Training Gopher}.
\newblock {\em arXiv preprint arXiv:2112.11446}, pages 1--120.

\bibitem[Raffel {\em et~al}., 2020]{c4_intro}
Raffel, C., Shazeer, N., Roberts, A., Lee, K., Narang, S., Matena, M., Zhou, Y., Li, W., and Liu, P.~J. (2020).
\newblock {Exploring the Limits of Transfer Learning with a Unified Text-to-text Transformer}.
\newblock {\em The Journal of Machine Learning Research}, 21(140):5485--5551.

\bibitem[Real {\em et~al}., 2020]{real2020assin}
Real, L., Fonseca, E., and Gon{\c{c}}alo~Oliveira, H. (2020).
\newblock {The ASSIN 2 Shared Task: A Quick Overview}.
\newblock In {\em Computational Processing of the Portuguese Language: 14th International Conference}, pages 406--412. Springer.

\bibitem[Roberts {\em et~al}., 2023]{roberts2023t5x}
Roberts, A., Chung, H.~W., Mishra, G., Levskaya, A., Bradbury, J., Andor, D., Narang, S., Lester, B., Gaffney, C., Mohiuddin, A., {\em et~al}. (2023).
\newblock {Scaling up models and data with t5x and seqio}.
\newblock {\em Journal of Machine Learning Research}, 24(377):1--8.

\bibitem[Rodrigues {\em et~al}., 2023]{albertina}
Rodrigues, J., Gomes, L., Silva, J., Branco, A., Santos, R., Cardoso, H.~L., and Osório, T. (2023).
\newblock {\em {Advancing Neural Encoding of Portuguese with Transformer Albertina PT-*}}, page 441–453.
\newblock Springer Nature Switzerland.
\newblock \url{http://dx.doi.org/10.1007/978-3-031-49008-8_35}.

\bibitem[Roziere {\em et~al}., 2023]{codellama}
Roziere, B., Gehring, J., Gloeckle, F., Sootla, S., Gat, I., Tan, X.~E., Adi, Y., Liu, J., Sauvestre, R., and Remez, T. (2023).
\newblock {Code Llama: Open Foundation Models for Code}.
\newblock {\em arXiv preprint arXiv:2308.12950}, pages 1--48.

\bibitem[Sakaguchi {\em et~al}., 2021]{sakaguchi2021winogrande}
Sakaguchi, K., Bras, R.~L., Bhagavatula, C., and Choi, Y. (2021).
\newblock {Winogrande: An Adversarial Winograd Schema Challenge at Scale}.
\newblock {\em Communications of the ACM}, 64(9):99--106.

\bibitem[Santos {\em et~al}., 2024]{gervasio}
Santos, R., Silva, J., Gomes, L., Rodrigues, J., and Branco, A. (2024).
\newblock {Advancing Generative AI for Portuguese with Open Decoder Gerv\'asio PT*}.
\newblock {\em arXiv 2402.18766}.
\newblock \url={https://arxiv.org/abs/2402.18766}.

\bibitem[Sayama {\em et~al}., 2019]{sayama2019faquad}
Sayama, H.~F., Araujo, A.~V., and Fernandes, E.~R. (2019).
\newblock {FaQuAD: Reading Comprehension Dataset in the Domain of Brazilian Higher Education}.
\newblock In {\em 8th Brazilian Conference on Intelligent Systems}, pages 443--448.

\bibitem[Shazeer and Stern, 2018]{shazeer2018adafactor}
Shazeer, N. and Stern, M. (2018).
\newblock {Adafactor: Adaptive Learning Rates with Sublinear Memory Cost}.
\newblock In {\em International Conference on Machine Learning}, pages 4596--4604. PMLR.

\bibitem[Silveira and Maua, 2018]{silveira2018enem}
Silveira, I.~C. and Maua, D.~D. (2018).
\newblock {Advances in Automatically Solving the ENEM}.
\newblock In {\em 7th Brazilian Conference on Intelligent Systems}, pages 43--48, Los Alamitos, CA, USA. IEEE Computer Society.

\bibitem[Socher {\em et~al}., 2013]{sst2}
Socher, R., Perelygin, A., Wu, J., Chuang, J., Manning, C.~D., Ng, A.~Y., and Potts, C. (2013).
\newblock {Recursive Deep Models for Semantic Compositionality over a Sentiment Treebank}.
\newblock In {\em Conference on Empirical Methods in Natural Language Processing}, pages 1631--1642.

\bibitem[Sorscher {\em et~al}., 2022]{pruning_1}
Sorscher, B., Geirhos, R., Shekhar, S., Ganguli, S., and Morcos, A. (2022).
\newblock {Beyond Neural Scaling Laws: Beating Power Law Scaling via Data Pruning}.
\newblock {\em Advances in Neural Information Processing Systems}, 35:19523--19536.

\bibitem[Souza {\em et~al}., 2020]{bertimbau}
Souza, F., Nogueira, R., and Lotufo, R. (2020).
\newblock {BERTimbau: Pretrained BERT Models for Brazilian Portuguese}.
\newblock In {\em Intelligent Systems: 9th Brazilian Conference}, pages 403--417. Springer.

\bibitem[Suarez {\em et~al}., 2020]{oscar_intro}
Suarez, P.~O., Romary, L., and Sagot, B. (2020).
\newblock {A Monolingual Approach to Contextualized Word Embeddings for Mid-Resource Languages}.
\newblock In {\em 58th Annual Meeting of the Association for Computational Linguistics}, pages 1703--1714.

\bibitem[Touvron {\em et~al}., 2023]{touvron2023llama2}
Touvron, H., Martin, L., Stone, K., Albert, P., Almahairi, A., Babaei, Y., Bashlykov, N., Batra, S., Bhargava, P., Bhosale, S., {\em et~al}. (2023).
\newblock {Llama 2: Open Foundation and Fine-Tuned Chat Models}.
\newblock {\em arXiv preprint arXiv:2307.09288}, pages 1--77.

\bibitem[Wei {\em et~al}., 2023]{magicoder}
Wei, Y., Wang, Z., Liu, J., Ding, Y., and Zhang, L. (2023).
\newblock {Magicoder: Source Code is All You Need}.
\newblock {\em arXiv preprint arXiv:2312.02120}, pages 1--26.

\bibitem[Wenzek {\em et~al}., 2019]{wenzek2019ccnet}
Wenzek, G., Lachaux, M.-A., Conneau, A., Chaudhary, V., Guzm{\'a}n, F., Joulin, A., and Grave, E. (2019).
\newblock {CCNet: Extracting High Quality Monolingual Datasets from Web Crawl Data}.
\newblock {\em arXiv preprint arXiv:1911.00359}.

\bibitem[Wu {\em et~al}., 2023]{bloomberggpt}
Wu, S., Irsoy, O., Lu, S., Dabravolski, V., Dredze, M., Gehrmann, S., Kambadur, P., Rosenberg, D., and Mann, G. (2023).
\newblock {BloombergGPT: A Large Language Model for Finance}.
\newblock {\em arXiv preprint arXiv:2303.17564}.

\bibitem[Xu {\em et~al}., 2024]{neuscraper}
Xu, Z., Liu, Z., Yan, Y., Liu, Z., Xiong, C., and Yu, G. (2024).
\newblock {Cleaner Pretraining Corpus Curation with Neural Web Scraping}.
\newblock {\em arXiv preprint arXiv:2402.14652}, pages 1--11.

\bibitem[Xue, 2020]{mc4}
Xue, L. (2020).
\newblock {mT5: A Massively Multilingual Pre-Trained Text-to-Text Transformer}.
\newblock {\em arXiv preprint arXiv:2010.11934}, pages 1--17.

\bibitem[Zhang {\em et~al}., 2024]{zhang2024tinyllama}
Zhang, P., Zeng, G., Wang, T., and Lu, W. (2024).
\newblock {TinyLlama: An Open-Source Small Language Model}.
\newblock {\em arXiv preprint arXiv:2401.02385}, pages 1--10.

\bibitem[Zhang {\em et~al}., 2015]{agnews}
Zhang, X., Zhao, J., and LeCun, Y. (2015).
\newblock {Character-level Convolutional Networks for Text Classification}.
\newblock In {\em 28th International Conference on Neural Information Processing Systems-Volume 1}, pages 649--657.

\end{thebibliography}

\appendix

\section{Rules Used for Document Selection}

In this appendix, we list all the rules from C4 and MassiveWeb used in our study.

Table~\ref{tab:massiveweb_filtering_rules} shows the MassiveWeb rules used in our study and the thresholds used. Table~\ref{tab:C4_used_rules} shows all the C4 rules used. As mentioned, we only used rules that did not involve partially editing a document. We also used the same list of Portuguese restricted words as mC4\footnote{\url{https://github.com/LDNOOBW/List-of-Dirty-Naughty-Obscene-and-Otherwise-Bad-Words/blob/master/pt}}.

\begin{table*}[!htb]
\setlength{\tabcolsep}{5mm}
\centering
\caption{Filtering rules applied from MassiveWeb~\citep{goupher_massiveweb}.}
\begin{tabular}{lc}
\toprule
\textbf{Description} & \textbf{Threshold} \\
\midrule
Number of words is too low & 50 \\
Number of words is too high & 100,000 \\
Mean word length is too short & 3 \\
Mean word length is too long & 10 \\
Symbol-to-word ratio (for \# or \ldots) is too high & 0.1 \\
Too many lines end with an ellipsis & 30\% \\
Too few words contain at least one alphabetic character & 90\% \\
Contains fewer than two stop words & 2 \\
\bottomrule
\end{tabular}
\label{tab:massiveweb_filtering_rules}
\end{table*}

\begin{table*}[!htb]
\setlength{\tabcolsep}{5mm}
\centering
\caption{Filtering rules applied from C4~\citep{c4_intro}.}
\label{tab:C4_used_rules}
\begin{tabular}{lc}
\toprule
\textbf{Rule Description}                     & \textbf{Threshold} \\ 
\midrule
Contains "\{"                                 & -                  \\ 
Document contains ``lorem ipsum''             & -                  \\ 
Document contains ``javascript''              & -                  \\ 
Document contains restricted words or phrases & -                  \\ 
Has fewer than the required sentences         & 3                  \\ 
\bottomrule
\end{tabular}
\end{table*}

\section{Full Results per Poeta Task}

\begin{figure*}[!htb]
\centering
\includegraphics[width=1\linewidth]{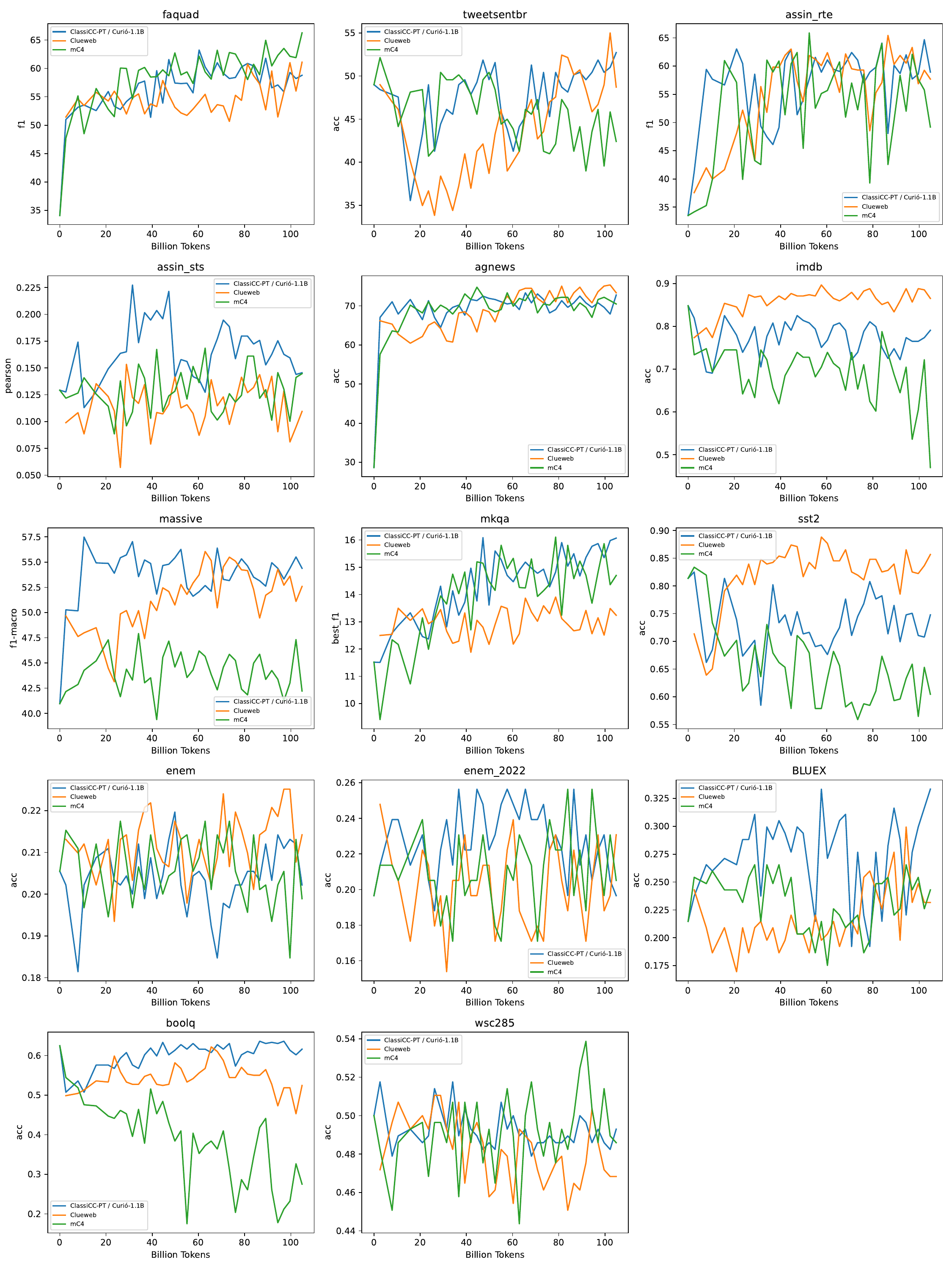}
\caption{Results for each task for the continued pretraining of TinyLlama 1T using mC4, ClueWeb and ClassiCC-PT.}
\label{fig:full_results_all_tasks}
\end{figure*}

Figure~\ref{fig:full_results_all_tasks} shows the performance of TinyLlama 1T's continued pretraining using mC4, ClueWeb, and ClassiCC-Pt for each task of the Poeta benchmark. 

We observed a unique behavior in the Agnews task: all continued pretraining led to a significant improvement early in training. This task requires the model to classify news articles into categories such as technology or sports. Initially, TinyLlama 1T, when prompted in Portuguese, almost always responded with 'Technology,' resulting in a random baseline score of around 25\%. After some training, this behavior disappeared, leading to the observed performance gain.

Furthermore, some tasks show little progress during training. In particular, all tasks based on Brazilian exams (ENEM and Bluex) exhibit performance only slightly above the expected random baseline.

\section{Prompts for Classifier Training}
\label{sec:appendixA}

This appendix provides the full prompts used for scoring documents in each category—educational, STEM, and toxic—for classifier training. Figure~\ref{fig:edu_classifier_prompt} shows the prompt for educational content, 
Figure~\ref{fig:stem_classifier_prompt} for STEM content, and Figure~\ref{fig:toxic_classifier_prompt} for toxic content.

\begin{figure}[!htb]
\centering
\begin{mdframed}[roundcorner=10pt]
    
Abaixo está um trecho de uma página da web. Avalie se a página tem um alto valor educacional e pode ser útil em um ambiente educacional para o ensino do ensino fundamental ao ensino médio usando o sistema de pontuação aditiva de 5 pontos descrito abaixo. Os pontos são acumulados com base na satisfação de cada critério:

- Adicione 1 ponto se o trecho fornecer algumas informações básicas relevantes para tópicos educacionais, mesmo que inclua algum conteúdo irrelevante ou não acadêmico, como anúncios e material promotional.

- Adicione outro ponto se o trecho abordar certos elementos pertinentes à educação, mas não alinhar-se de perto com os padrões educacionais. Pode misturar conteúdo educacional com material não educacional, oferecendo uma visão geral superficial de tópicos potencialmente úteis ou apresentando informações de maneira desorganizada e estilo de escrita incoerente.

- Conceda um terceiro ponto se o trecho for apropriado para uso educacional e introduzir conceitos-chave relevantes para o currículo escolar. É coerente, embora possa não ser abrangente ou possa incluir algumas informações extrínsecas. Pode assemelhar-se a uma seção introdutória de um livro didático ou a um tutorial básico que seja adequado para aprendizado, mas tem limitações notáveis, como tratar de conceitos que são muito complexos para alunos do ensino fundamental.

- Conceda um quarto ponto se o trecho for altamente relevante e benéfico para fins educacionais em um nível não superior ao ensino fundamental, exibindo um estilo de escrita claro e consistente. Pode ser semelhante a um capítulo de um livro didático ou um tutorial, oferecendo conteúdo educacional substancial, incluindo exercícios e soluções, com informações mínimas irrelevantes, e os conceitos não são muito avançados para alunos do ensino fundamental. O conteúdo é coerente, focado e valioso para aprendizado estruturado.

- Conceda um quinto ponto se o trecho for excepcional em seu valor educacional, perfeitamente adequado para ensino no ensino fundamental ou médio. Segue um raciocínio detalhado, o estilo de escrita é fácil de seguir e oferece insights profundos e completos sobre o assunto, desprovidos de qualquer conteúdo não educacional ou complexo.

O trecho: \{text\}.

Após examinar o trecho:

    Justifique brevemente sua pontuação total, até 100 palavras.
    Conclua com a pontuação usando o formato: "Pontuação educacional: <total de pontos>
    \end{mdframed}
    \caption{Education prompt for the labeling of data used to create the ClassiCC-PT-Edu Classifier.}
    \label{fig:edu_classifier_prompt}
\end{figure}

\begin{figure}[!htb]
\centering
\begin{mdframed}[roundcorner=10pt]
Abaixo está um trecho de uma página da web. Avalie o quão bem o conteúdo do trecho está alinhado com temas de STEM e sua adequação para um público interessado em aprimorar conhecimentos nessas áreas, utilizando um sistema de pontuação aditiva de 5 pontos. Os pontos são acumulados com base na satisfação de cada critério:

- Adicione 1 ponto se o trecho mencionar conceitos de STEM, mesmo que de forma superficial ou tangencial, e independentemente de conter conteúdo promocional ou publicitário.

- Adicione outro ponto se o trecho discutir temas de STEM de maneira introdutória, mas com alguma falta de profundidade ou precisão técnica. Pode incluir exemplos ou aplicações práticas simplificadas.

- Conceda um terceiro ponto se o trecho apresentar informações de STEM que são factualmente corretas e relevantes para a área, mas ainda de maneira não muito detalhada. Pode ser um bom ponto de partida para discussões em salas de aula de nível técnico ou universitário.

- Conceda um quarto ponto se o trecho oferecer uma explicação detalhada de um conceito de STEM, com exemplos práticos e relevância demonstrada para aplicações reais, sendo útil para estudantes de nível técnico ou universitário.

- Conceda um quinto ponto se o trecho for uma referência excelente e detalhada em STEM, com análise técnica profunda, exemplos práticos avançados e relevância direta para campos profissionais ou estudos avançados em STEM.

O trecho: \{text\}.

Após examinar o trecho:

    Justifique brevemente sua pontuação total, até 100 palavras.
    Conclua com a pontuação usando o formato: "Pontuação STEM: <total de pontos>"
    \end{mdframed}
    \caption{Education erompt for the labeling of data used to create the ClassiCC-PT-STEM Classifier.}
    \label{fig:stem_classifier_prompt}
\end{figure}

\begin{figure}[!htb]
\centering
\begin{mdframed}[roundcorner=10pt]
Abaixo está um trecho de uma página da web. Avalie se a página contém conteúdo ofensivo, incluindo, mas não se limitando a, material pornográfico, discurso de ódio ou retórica extremista, utilizando o sistema de pontuação aditiva de 5 pontos descrito abaixo. Os pontos são acumulados com base na presença e gravidade de cada critério:

    - Adicione 1 ponto se o trecho contiver sugestões leves de conteúdo ofensivo, como linguagem imprópria ou referências subtis a temas adultos.
    
    - Adicione outro ponto se o trecho incluir conteúdo que possa ser considerado ofensivo por algumas pessoas, como humor ácido ou discussões controversas, mas sem material explícito.
    
    - Conceda um terceiro ponto se o trecho apresentar conteúdo que seja claramente ofensivo para muitos, como representações estereotipadas ou piadas de mau gosto, mas ainda não atinge o nível de material explícito.
    
    - Conceda um quarto ponto se o trecho for predominantemente ofensivo, com linguagem e temas que são considerados inapropriados para a maioria dos públicos, mas não contém material explícito de natureza pornográfica, violenta ou extremista.
    
    - Conceda um quinto ponto se o trecho contiver conteúdo explícito que é pornográfico, constitui discurso de ódio ou promove retórica extremista, tornando-o altamente ofensivo e inapropriado.

O trecho: \{text\}.

Após examinar o trecho:

    Justifique brevemente sua pontuação total, até 100 palavras.
    Conclua com a pontuação usando o formato: "Pontuação ofensiva: <total de pontos>".
    \end{mdframed}
    \caption{Education prompt for the labeling of data used to create the ClassiCC-PT-Toxic Classifier.}
    \label{fig:toxic_classifier_prompt}
\end{figure}

\section{Contamination Analysis}

We performed a contamination analysis of the main datasets used in this study—namely, ClassiCC, ClueWeb-A, and mC4-PT. Following the methodology developed by~\citet{achiam2023gpt4_technical}, we consider a document contaminated if three 50-character substrings from an evaluation example are found in a training document.

Using this method, we found considerable contamination only in two Poeta datasets (ENEM and BLUEX), expected given that these benchmarks cover prominent entrance exams in Brazil. We also identified contaminated examples in all three pretraining datasets.

\begin{figure}[!htb]
    \centering
    \includegraphics[width=1\linewidth]{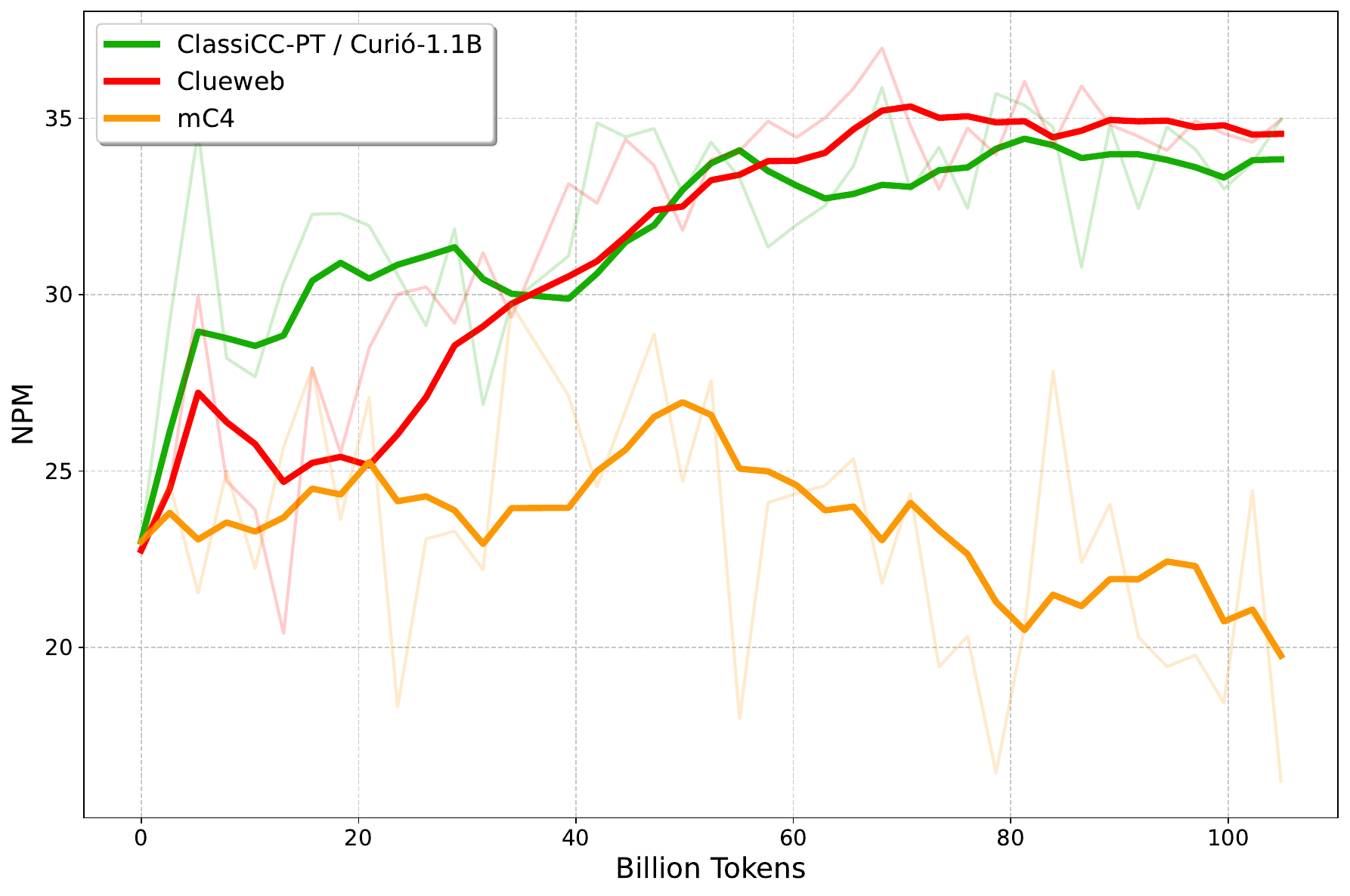}
    \caption{Average NPM for non-contaminated tasks.}
    \label{fig:npm_no_exmas}
\end{figure}

However, since our experiments were limited to models at the 1.1B scale, none of the trained models achieved performance significantly above random chance on these benchmarks. Figure~\ref{fig:npm_no_exmas} shows the models' performance excluding all tasks with considerable contamination. The results are very similar to those in Figure~\ref{fig:full_results_all_tasks}, indicating that contamination likely did not have a significant impact on our evaluations.

\section{Distribution of LLM Annotated Documents}
\label{appendix:score_dist}

As mentioned previously, we used GPT-4o to annotate hundreds of thousands of documents across three categories: Educational, STEM, and Toxic. The model assigns a score between 0 and 5, based on the prompts shown in Appendix~\ref{sec:appendixA}.

Table~\ref{tab:dist_scores} presents the distribution of scores assigned by GPT-4o across each category. In all cases, we observe a higher concentration of lower scores, which is expected since our sample consists of randomly selected web documents. Many of these pages are, for example, simple listings or advertisements that do not correspond to any of the evaluated categories.

\begin{table*}[!htb]
\setlength{\tabcolsep}{5mm}
\centering
\caption{Score distribution of the GPT-4o annotation of documents though all three categories.}
\label{tab:dist_scores}
\begin{tabular}{@{}lcccccc@{}}
\toprule
\textbf{Scores}      & \textbf{0}    & \textbf{1}    & \textbf{2}    & \textbf{3}    & \textbf{4}   & \textbf{5}   \\ \midrule
Toxic       & 94\% & 2\%  & 1\%  & 1\%  & 1\% & 1\% \\
Stem        & 67\% & 17\% & 4\%  & 10\% & 1\% & 1\% \\
Educational & 52\% & 24\% & 14\% & 8\%  & 1\% & 1\% \\ \bottomrule
\end{tabular}
\end{table*}

\end{document}